\documentclass[10pt,twocolumn,letterpaper]{article}

\usepackage{3dv}

\usepackage{graphicx}
\usepackage{amsmath,amssymb} 
\usepackage{color}

\usepackage{arydshln}
\usepackage[dvipsnames]{xcolor}
\usepackage{soul}
\usepackage{url}
\usepackage{booktabs}
\usepackage{multirow}
\usepackage{float}
\usepackage{dblfloatfix}
\usepackage{wrapfig}
\usepackage{algorithm}
\usepackage{capt-of}
\usepackage[noend]{algpseudocode}
\newcommand\blfootnote[1]{%
  \begingroup
  \renewcommand\thefootnote{}\footnote{#1}%
  \addtocounter{footnote}{-1}%
  \endgroup
}
\usepackage[breaklinks=true,bookmarks=false]{hyperref}

\newcommand\T{\rule{0pt}{2.6ex}}       
\newcommand{\change}[1]{#1}

\newcommand{\parahead}[1]{\textbf{#1}:\ }

\newenvironment{packed_itemize}
{\begin{itemize}
    \setlength{\itemsep}{1pt}
    \setlength{\parskip}{0pt}
    \setlength{\parsep}{0pt}
}{\end{itemize}}

\threedvfinalcopy 

\ifthreedvfinal\fi
\begin{document}
\title{Single-Shot Multi-Person 3D Pose Estimation From Monocular RGB}

\author{Dushyant Mehta$^{[1,2]}$, Oleksandr Sotnychenko$^{[1,2]}$, Franziska Mueller$^{[1,2]}$,\\Weipeng Xu$^{[1,2]}$, Srinath Sridhar$^{[3]}$, Gerard Pons-Moll$^{[1,2]}$, Christian Theobalt$^{[1,2]}$
\and 
[1] MPI For Informatics   \: [2] Saarland Informatics Campus  \: [3] Stanford University \vspace{-0.5cm}}

\maketitle

\begin{figure*}
\centering
  \includegraphics[width=\linewidth]{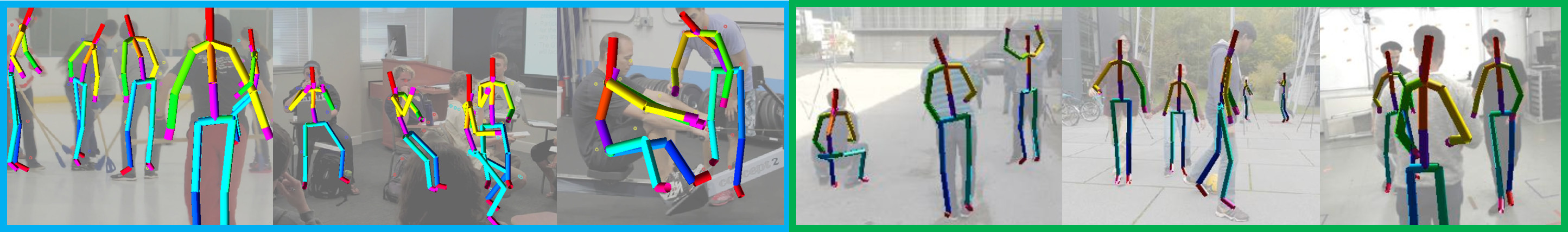}
  \vspace{-0.42cm}
  \caption
  { Qualitative results of our approach shown on MPII 2D~\cite{andriluka_mpii2d_cvpr14} dataset (blue), as well as our new \emph{MuPoTS-3D} evaluation set (green). Our pose estimation approach works for general scenes, handling occlusions by objects or other people. Note that our 3D pose predictions are root-relative, and scaled and overlaid only for visualization.}   
  \vspace{-0.5cm}
  \label{fig:teaser}
\end{figure*}
\begin{abstract}
We propose a new single-shot method for multi-person 3D pose estimation in general scenes from a monocular RGB camera. 
Our approach uses novel occlusion-robust pose-maps (ORPM) which enable full body pose inference even under strong partial occlusions by other people and objects in the scene.
ORPM outputs a fixed number of maps which encode the 3D joint locations of all people in the scene.
Body part associations \cite{cao_affinity_2017} allow us to infer 3D pose for an arbitrary number of people without explicit bounding box prediction.
To train our approach we introduce \emph{MuCo-3DHP}, the first large scale training data set showing real images of sophisticated multi-person interactions and occlusions.
We synthesize a large corpus of multi-person images by compositing images of individual people (with ground truth from mutli-view performance capture).
We evaluate our method on our new challenging 3D annotated multi-person test set \emph{MuPoTs-3D} where we achieve state-of-the-art performance.
To further stimulate research in multi-person 3D pose estimation, we will make our new datasets, 
and associated code publicly available for research purposes.
\end{abstract}
\blfootnote{This work was funded by the ERC Starting Grant project CapReal (335545). We would also like to thank Hyeongwoo Kim and Eldar Insafutdinov for their assistance.}
\section{Introduction}
\label{sec:intro}
Single-person pose estimation, both 2D and 3D, from monocular RGB input is a challenging and widely studied problem in vision~\cite{andriluka_pictorial_cvpr09,andriluka_mpii2d_cvpr14,mehta_mono_3dv17,VNect_SIGGRAPH2017,bulat_convpart_eccv16,chen_nips14,li_maximum_iccv2015,newell_stacked_hourglass_eccv16}. It has many applications, \eg, in activity recognition and content creation for graphics.
While methods for 2D multi-person pose estimation exist~\cite{pishchulin_deepcut_cvpr16,insafutdinov_deepercut_eccv16,cao_affinity_2017,newell_stacked_hourglass_eccv16}, most 3D pose estimation methods are restricted to a single un-occluded subject. 
Natural human activities take place with multiple people in cluttered scenes hence exhibiting not only self-occlusions of the body, but also strong inter-person occlusions or occlusions by objects.
This makes the under-constrained problem of inferring 3D pose (of all subjects) from monocular RGB input even harder and leads to drastic failure of existing single person 3D pose estimation methods.

Recent work approaches this more general 3D multi-person pose estimation problem by decomposing it into multiple single-person instances~\cite{luvizon20182d,rogez_lcr_cvpr17}, often with significant redundancy in the decomposition~\cite{rogez_lcr_cvpr17}. The single person predictions are post-processed to filter, refine and fuse the predictions into a coherent estimate. Bottom-up joint multi-person reasoning remains largely unsolved, and the multi-person 3D pose estimation lacks appropriate performance benchmarks. 

We propose a new \emph{single shot} CNN-based method to estimate multi-person 3D pose in general scenes from monocular input. 
We call our method single shot since it reasons about all people in a scene jointly in a single forward pass, and does not require explicit bounding box proposals by a separate algorithm as a preprocessing step~\cite{rogez_lcr_cvpr17,papandreou2017towards}. 
The latter may fail under strong occlusions and may be expensive to compute in dense multi-person scenes.
Our fully-convolutional method jointly infers 2D and 3D joint locations using our new occlusion-robust pose-map (ORPM) formulation.
ORPM enables multi-person 3D pose estimates under strong (self-)occlusions by incorporating redundancy in the encoding, while using a fixed number of outputs regardless of the number of people in the scene.
Our subsequent hierarchical read-out strategy starts with a base pose estimate, and is able to refine the estimate based on which joints of a person are visible, leading to robust 3D pose results.

To train our CNN we introduce a new multi-person 3D pose data set \emph{MuCo-3DHP}. 
While there are several single-person datasets with 3D joint annotations, there are no annotated multi-person datasets containing large corpora of real video recordings of human--human interaction with large person and background diversity.
Important advances in this direction have been made by Joo \etal~\cite{joo_panoptic_iccv2015} using a multi-camera studio setup but background diversity remains limited.
Some prior work creates 3D annotated multi-person images by using 2D pose data augmented with 3D poses from motion capture datasets~\cite{rogez_lcr_cvpr17}, or by finding 3D consistency in 2D part annotations from multi-view images recorded in a studio~\cite{simon2017hand}.
To create training data with much larger diversity in person appearance, camera view, occlusion and background, we transform the MPI-INF-3DHP single-person dataset~\cite{mehta_mono_3dv17} into the first multi-person set that shows images of real people in complex scenes.
\emph{MuCo-3DHP} is created by compositing multiple 2D person images with ground-truth 3D pose from multi-view marker-less motion capture. 
Background augmentation and shading-aware foreground augmentation of person appearance enable further data diversity.
To validate the generalizability of our approach to real scenes, and since there are only very few annotated multi-person test sets~\cite{elhayek_convmocap_TPAMI2016} showing more than two people, we contribute a new multi-person 3D test set, \emph{MuPoTS-3D}.
It features indoor and outdoor scenes, challenging occlusions and interactions, varying backgrounds, more than two persons, and ground truth from commercial marker-less motion capture. All datasets will be made publicly available. 
In summary, we contribute:
\begin{packed_itemize}
\item A CNN-based single-shot \emph{multi-person pose estimation method} based on a novel multi-person 3D pose-map formulation to jointly predict 2D and 3D joint locations of all persons in the scene. 
Our method is tailored for scenes with occlusion by objects or other people. 
\item The first \emph{multi-person dataset} of real person images with 3D ground truth that contains complex inter-person occlusions, motion, and background diversity. 
\item A real \emph{in-the-wild test set} for evaluating multi-person 3D pose estimation methods that contains diverse scenes, challenging multi-person interactions, occlusions, and motion.
\end{packed_itemize}

Our method achieves state-of-the-art performance on challenging multi-person scenes where single-person methods completely fail. Although designed for the much harder multi-person task, it performs competitively on single-person test data.
\section{Related Work} 
\label{sec:rel-work}
We focus on most directly related work estimating the pose of \emph{multiple people in 2D} or a \emph{single person in 3D} from  monocular RGB input.
\cite{sarafianos_posesurvey_cviu2016} provide a more comprehensive review.

\parahead{Multi-Person 2D Pose Estimation}
A common approach for multi-person 2D pose estimation is to first detect single persons and then 2D pose~\cite{pishchulin_reshape_cvpr12,gkioxari2014using,sun2011articulated,iqbal2016multi,papandreou2017towards}.
Unfortunately, these methods fail when the detectors fail---a likely scenario with multiple persons and strong occlusions. 
Hence, a body of work first localizes the joints of each person with CNN-based detectors and then find the correct association between joints and subjects in a post-processing step~\cite{pishchulin_deepcut_cvpr16,insafutdin_arttrack_cvpr17,newell_associative_nips17,cao_affinity_2017}. 

\parahead{Single-Person 3D Pose Estimation}
Existing monocular single-person 3D pose methods show good performance on standard datasets~\cite{ionescu_human36_pami14,sigal_humaneva_ijcv10,tekin_structured_bmvc16,pavlakos_volumetric_cvpr17,li_maximum_iccv2015,li_accv14}. 
However, since many methods train a discriminative predictor for 3D poses~\cite{bo_twin_ijcv10}, they often do not generalize well to natural scenes with varied poses, appearances, backgrounds and occlusions. 
This is due to the fact that most 3D datasets are restricted to indoor setups with limited backgrounds and appearance.
The advent of large real world image datasets with 2D annotations made 2D pose estimation in the wild remarkably accurate.
However, annotating images with 3D pose is much harder with many recent work focusing on leveraging 2D image datasets for 3D human pose estimation or multi-view settings~\cite{joo_panoptic_iccv2015}. Additional annotations to these 2D image datasets allow some degree of 3D reasoning, either through body joint depth ordering constraints~\cite{pavlakos2018humanshape} or dense shape correspondences~\cite{guler2018densepose}.
Some works split the problem in two: first estimate 2D joints and then lift them to 3D~\cite{taylor_articulated_cvpr00,sminchisescu2003kinematic,chen_2d_match_cvpr17,yasin_dual_source_cvpr16,moreno_distance_matrix_cvpr17,zhou_sparseness_deepness_cvpr15,akhter_pose_conditioned_cvpr15,simo_joint_CVPR2013,Jahangiri2017,martinez20173dbaseline,bogo_smpl_eccv16,Lassner:UP:2017,nie2017monocular,tan2017indirect,alldieck2018video}, \eg, by database matching, neural network regression, or fitting the SMPL body model~\cite{SMPL:2015}. Some works integrate SMPL within the CNN to exploit 3D and 2D annotations in an end-to-end fashion~\cite{omran2018neural,hmrKanazawa17,pavlakos2018humanshape}.

Other work leverages the features learned by a 2D pose estimation CNN for 3D pose estimation.
For example, \cite{tekin_fusion_arxiv16} learn to merge features from a 2D and 3D joint prediction network.
Another approach is to train a network with separate 2D and 3D losses for the different data sources~\cite{popa2017deep,zhou2017towards,sun2017compositional,yang20183d,luvizon20182d}. Some approaches jointly reason about 2D and 3D pose with multi-stage belief maps~\cite{tome_lifting_2017}. 
The advantage of such methods is that they can be trained end to end. 
A simpler yet very effective approach is to refine a network trained for 2D pose estimation for the task of 3D pose estimation~\cite{VNect_SIGGRAPH2017,mehta_mono_3dv17}. 
A major limitation of methods that rely on 2D joint detections directly or on bounding boxes is that they easily fail under body occlusion or with incorrect 2D detections, both of which are common in multi-person scenes. 
In contrast, our approach is more robust to occlusions since a base 3D body pose estimate is available even under significant occlusion. 
\cite{VNect_SIGGRAPH2017} showed that 3D joint prediction works best when the receptive field is centered around the joint of interest.
We build upon this insight to refine the base body pose where 2D joint detections are available.

\begin{figure}
\centering
\centering
  \includegraphics[width=0.95\linewidth]{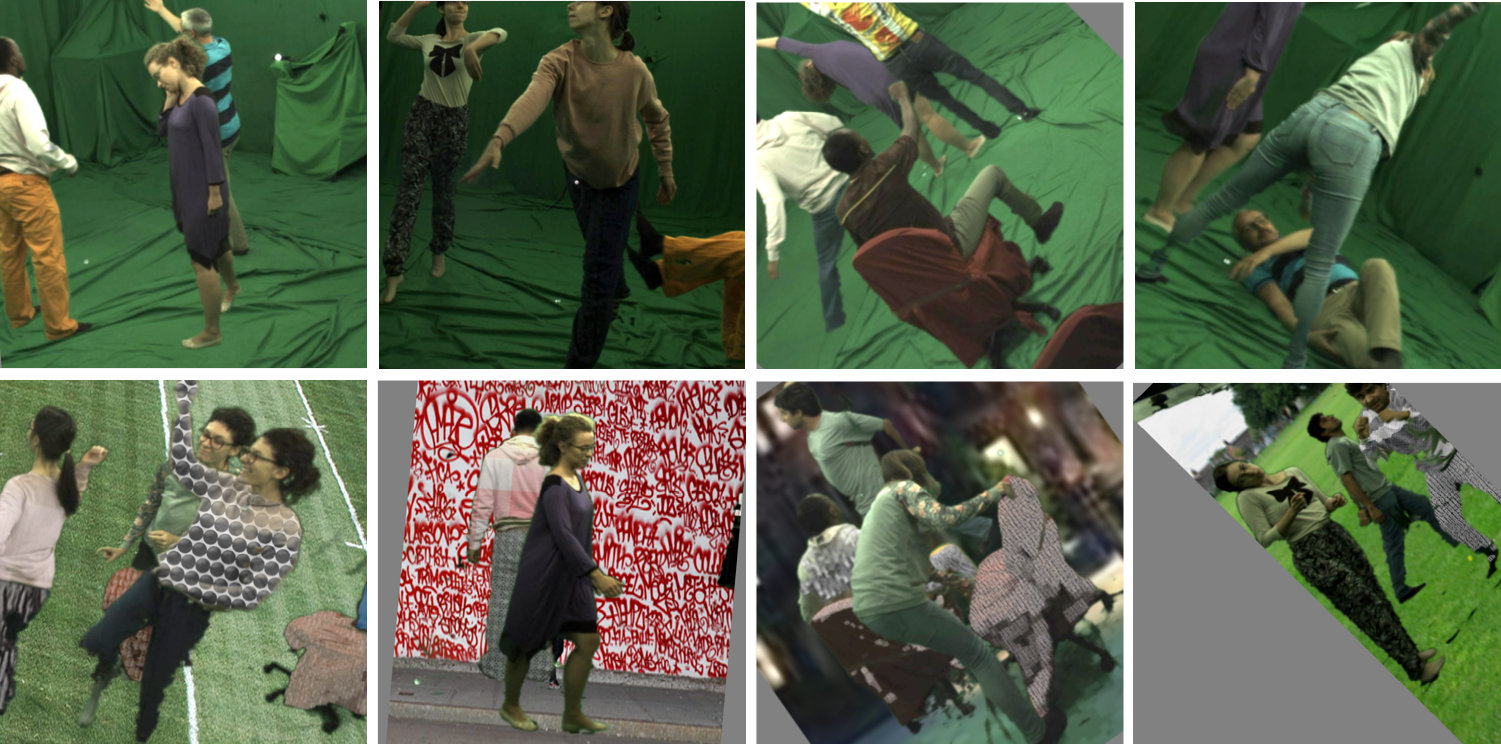}
  \vspace{-0.0cm}
  \caption
  {
  Examples from our \textbf{MuCo-3DHP} dataset, created through compositing MPI-INF-3DHP~\cite{mehta_mono_3dv17} data. (Top) composited examples without appearance augmentation, (bottom) with BG and clothing augmentation. The last two columns show rotation and scale augmentation, and truncation with the frame boundary.
  }  
  \label{fig:training_set}
  \vspace{-0.5cm}
\end{figure}
\parahead{Multi-Person 3D Pose Estimation}
To our knowledge, only \cite{rogez_lcr_cvpr17} tackle multi-person 3D pose estimation from single images.\footnote{A second approach~\cite{zanfir2018monocular} was published in the review period.}
%
They first identify bounding boxes likely to contain a person using~\cite{ren_faster_rcnn_nips15}.
Instead of a direct regression to pose, the bounding boxes are classified into a set of K-poses similar to~\cite{posebits_cvpr14}. 
These poses are scored by a classifier and refined using a regressor.
The method implicitly reasons using bounding boxes and produces multiple proposals per subject that need to be accumulated and fused. 
However, performance of their method under large person-person occlusions is unclear. 
In contrast, our approach produces multi-person 2D joint locations and 3D pose maps in a single shot, from which the 3D pose can be inferred even under severe person-person occlusion.

\parahead{3D Pose Datasets}
Existing pose datasets are either for a single person in 3D~\cite{ionescu_human36_pami14,sigal_humaneva_ijcv10,trumble2017total,vonPon2016a,mehta_mono_3dv17} or multi-person with only 2D pose annotations~\cite{andriluka_mpii2d_cvpr14,lin_coco_eccv14}. 
Of the two exceptions, the MARCOnI dataset~\cite{elhayek_convmocap_TPAMI2016} features 5 sequences but contains only 2 persons simultaneously, and there are no close interactions.
The other is the Panoptic dataset~\cite{joo_panoptic_iccv2015} which has a limited capture volume, pose and background diversity. 
There is work on generating synthetic images~\cite{rogez_mocap_nips16,chen_synth_data_3dv16} from mocap data, however the resulting images are not plausible.
%
We choose to leverage the person segmentation masks available in MPI-INF-3DHP~\cite{mehta_mono_3dv17} to generate annotated multi-person 3D pose images of real people at scale through compositing using the available segmentation masks.
Furthermore, we captured a 3D benchmark dataset featuring multiple closely interacting persons which was annotated by a video-based multi-camera motion capture system.

\section{Multi-Person Dataset}
\label{sec:dataset}
Generating data by combining in-the-wild multi-person 2D pose data~\cite{andriluka_mpii2d_cvpr14,lin_coco_eccv14} and multi-person multi-view motion capture for 3D annotation would be a straightforward extension of previous (single-person) approaches~\cite{mehta_mono_3dv17}. 
However, multi-person 3D motion capture under strong occlusions and interactions is challenging even for commercial systems, often requiring manual pose correction constraining 3D accuracy. 
%
%
Hence, we merely employ purely multi-view marker-less motion capture to create the 20 sequences of \emph{MuPoTs-3D}, the first expressive in-the-wild multi-person 3D pose benchmark.
For the much larger training set \emph{MuCo-3DHP}, we resort to a new compositing and augmentation scheme that leverages the single-person image data of real people in MPI-INF-3DHP\cite{mehta_mono_3dv17} to composite an arbitrary number of multi-person interaction images under user control, with 3D pose annotations.

\subsection{MuCo-3DHP: Compositing-Based Training Set}
The MPI-INF-3DHP~\cite{mehta_mono_3dv17} single-person 3D pose dataset provides marker-less motion capture based annotations for real images of 8 subjects, each captured with 2 clothing sets, using 14 cameras at different ele\-vations. 
We build upon these person segmentation masks to create per-camera composites with 1 to 4 subjects, with frames randomly selected from the $8\times2$ sequences available per camera. 
Since we have ground-truth 3D skeleton pose for each video subject in the same space, we can composite in a 3D-aware manner resulting in correct depth ordering and overlap of subjects.
We refer to this composited training set as the \textbf{Mu}ltiperson \textbf{Co}mposited \textbf{3D} \textbf{H}uman \textbf{P}ose dataset (see Fig.~\ref{fig:training_set} for examples). 
The compositing process results in plausible images covering a range of simulated inter-person overlap and activity scenarios. 
Furthermore, user-control over the desired pose and occlusion distribution, and foreground/background augmentation using the masks provided with MPI-INF-3DHP is possible (see supplementary document for more details). 
%
Even though the synthesized composites may not simulate all the nuances of human-human interaction fully, we observe that our approach trained on this data generalizes well to real world scenes in the test set. 
 
\begin{figure}
\centering
  \includegraphics[width=0.95\linewidth]{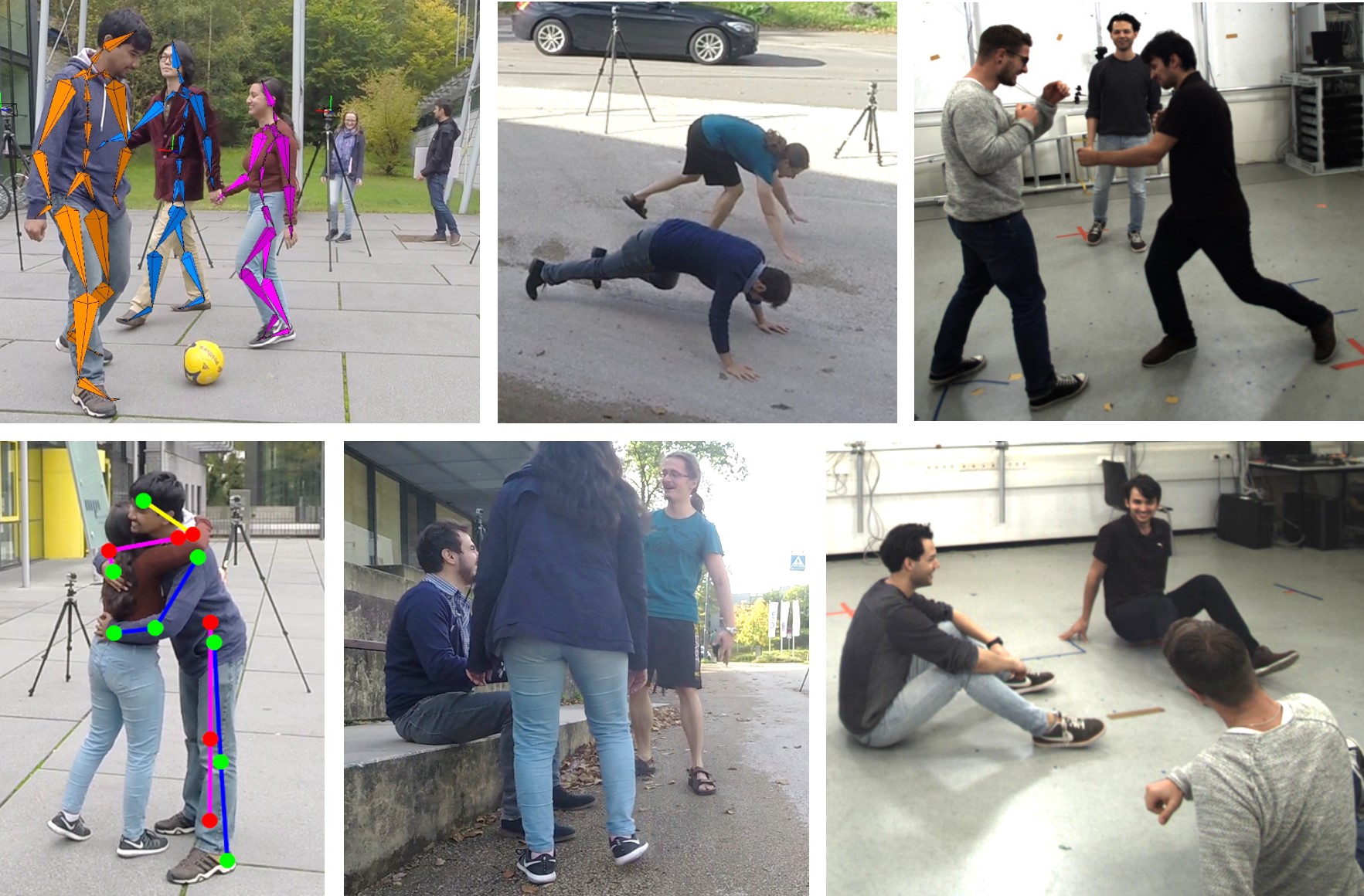}
  \vspace{-0.0cm}
  \caption
  {
  Examples from our \emph{MuPoTS-3D} evaluation set. Ground truth 3D pose reference and joint occlusion annotations are available for up to 3 subjects in the scene. The set covers a variety of scene settings, activities and clothing.
  }  
  \label{fig:test_set}
  \vspace{-0.6cm}
\end{figure}
\subsection{MuPoTS-3D: Diverse Multi-Person 3D Test Set}
\label{sec:test_set}
We also present a new filmed (not composited) multi-person test set comprising 20 general real world scenes with ground-truth 3D pose for up to three subjects obtained with a multi-view marker-less motion capture system~\cite{Captury}. 
Additionally, per joint occlusion annotations are available.
The set covers 5 indoor and 15 outdoor settings, with trees, office buildings, road, people, vehicles, and other stationary and moving distractors in the background. 
Some of the outdoor footage also has challenging elements like drastic illumination changes, and lens flare. 
The indoor sequences use $2048\times2048$px footage at 30fps, and outdoor sequences use $1920\times1080$px GoPro footage at 60fps.
The test set consists of $>$8000 frames, split among the 20 sequences, with 8 subjects, in a variety of clothing styles, poses, interactions, and activities. 
Notably, the test sequences do not resemble the training data, and include real interaction scenarios.
We call our new test set \textbf{Mu}ltiperson \textbf{Po}se \textbf{T}est \textbf{S}et in \textbf{3D} (\emph{MuPoTS-3D}).

\parahead{Evaluation Metric}
We use the robust \emph{3DPCK} evaluation metric proposed in~\cite{mehta_mono_3dv17}. It treats a joint's prediction as correct if it lies within a 15cm ball centered at the ground-truth joint location, and is evaluated for the common minimum set of 14 joints marked in green in Fig.~\ref{fig:pose_representation}. 
We report the \emph{3DPCK} numbers per sequence, averaged over the subjects for which GT reference is available, \change{and additionally report the performance breakdown for occluded and un-occluded joints.}
\change{The relative robustness of 3DPCK over MPJPE\cite{ionescu_human36_pami14} is also useful to offset the effect of jitter that arises in all non-synthetic annotations, including ours.} For completeness, we also report the MPJPE error for predictions matched to an annotated subject.
\newcommand{\newLocationMap}{ORPM~}
\newcommand{\newLocationMaps}{ORPMs~}
\section{Method}
\label{sec:method}
At the core of our approach is a novel formulation which allows us to estimate the pose of multiple people in a scene even under strong occlusions with a single forward pass of a fully convolutional network.
Our method builds upon the \emph{location-maps} formulation~\cite{VNect_SIGGRAPH2017} that links 3D pose inference more strongly to image evidence by inferring 3D joint positions at the respective 2D joint pixel locations.
%
%
%
%
We first recap the location-map formulation before describing our approach.

\begin{figure}[t]
  \begin{center}
  \includegraphics[width=0.95\linewidth]{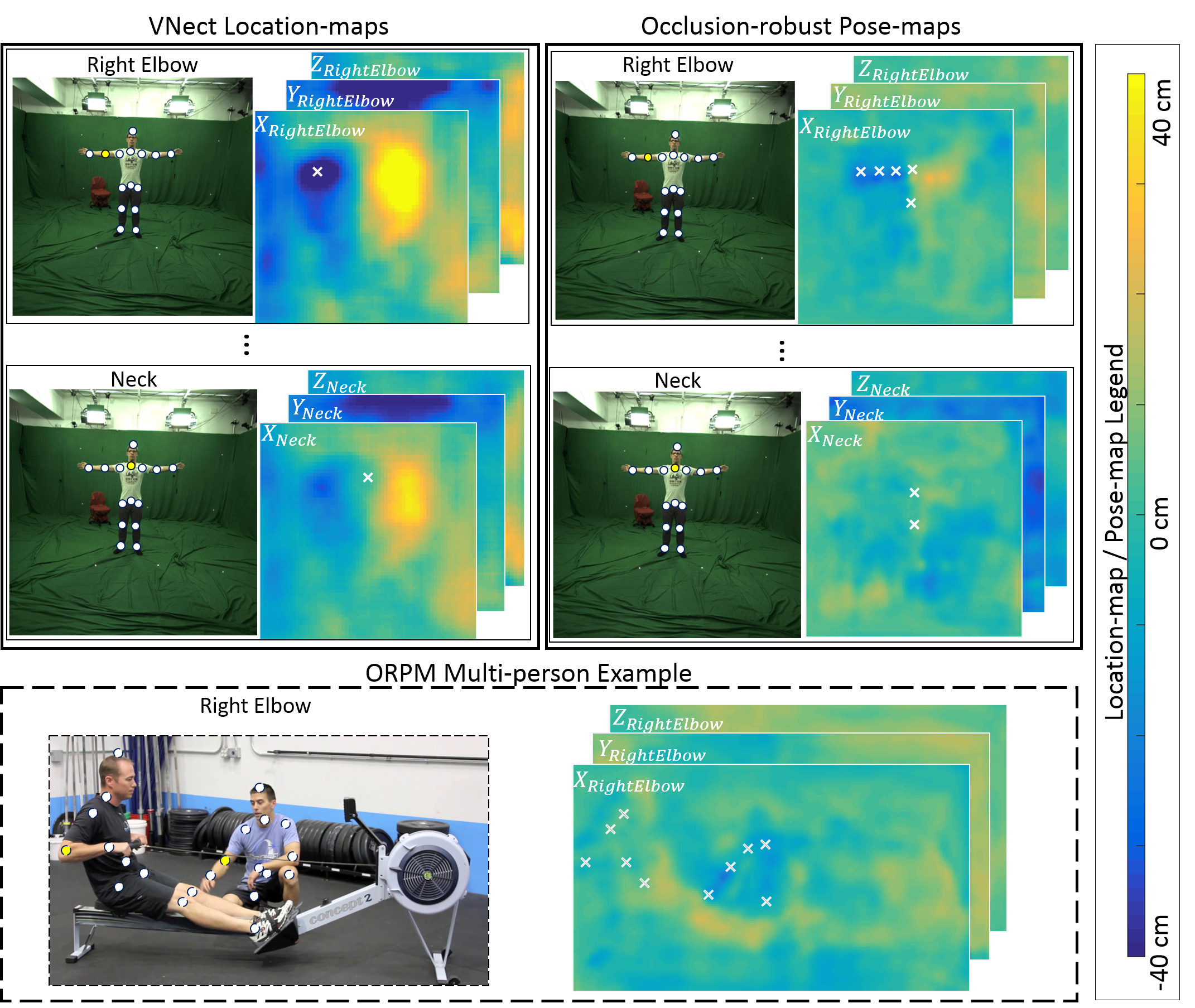}
  \end{center}
  \vspace{-0.3cm}
  \caption
  {
  Multiple levels of selective redundancy in our Occlusion-robust Pose-map (ORPM) formulation. VNect Location-maps~\cite{VNect_SIGGRAPH2017} (left) only support readout at a single pixel location per joint type. ORPMs (middle) allow the complete body pose to be read out at torso joint pixel locations (neck, pelvis). Further, each individual limb's pose can be read out at all 2D joint pixel locations of the respective limb. This translates to read-out of each joint's location being possible at multiple pixel locations in the joint's location map. 
  The example at the bottom shows how 3D locations of multiple people are encoded into the same map per joint and no additional channels are required.
  }  
  \vspace{-0.4cm}
  \label{fig:training_location_maps}
\end{figure}

\parahead{Location-Maps~\cite{VNect_SIGGRAPH2017}}
A location-map is a joint specific feature channel storing the 3D coordinate $x$, $y$, or $z$ at the joint 2D pixel location.
For every joint, three location-maps, as well as a 2D pixel location heatmap are estimated. 
The latter encodes the 2D pixel location of the joint as a confidence map in the image plane. The 3D position of a joint can be read out from its location-map at the 2D pixel location of the joint, as shown in Fig.~\ref{fig:training_location_maps}.
For an image of size $W\times{H}$, $3n$ location-maps of size ${W/k}\times{H/k}$ are used to store the 3D location of all $n$ joints, where $k$ is a down-sampling factor.
During training, the $L_2$ loss between the ground truth and the estimated location-map is minimized in the area around the joint's 2D pixel location.  
Although this simple location-map formulation enables full 3D pose inference, it has several shortcomings.
First, it assumes that all joints of a person are fully visible, and breaks down under partial occlusion, which is common in general scenes.
Second, efficient extension to multiple people is not straightforward.
Introducing separate location-maps per person requires dynamically changing the number of outputs.

\parahead{Occlusion-Robust Pose-Maps (ORPMs)}
We propose a novel occlusion-robust formulation that has a fixed number of outputs regardless of the number of people in the scene, while enabling pose read-outs for strongly occluded people.
Our key insight is the incorporation of \emph{redundancy} into the location-maps. 
We represent the body by decomposing it into torso, four limbs, and head (see Fig.~\ref{fig:pose_representation}).
Our occlusion-robust pose-maps (ORPMs) support multiple levels of redundancy: 
(1)~they allow the read-out of the complete \emph{base pose} $\mathbf{P}\in \mathbb{R}^{3 \times n}$ at one of the torso joint locations (neck or pelvis),
(2)~the base pose (which may not capture the full extent of articulation) can be further refined by reading out the head and individual limb poses where 2D detections are available,
and (3)~the complete limb pose can be read out at any 2D joint location of that limb.
Together, these ensure that a complete and as articulate as possible pose estimate is available even in the presence of heavy occlusions of the body (see Fig.~\ref{fig:pose_representation}, and Fig.~2,3 in the supplementary document).
In addition, the redundancy in \newLocationMaps allows to encode the pose of multiple partially overlapping persons without loss of information, thus removing the need for a variable number of output channels. See Fig.~\ref{fig:training_location_maps}.

\parahead{Na\"ive Redundancy}
The na\"ive approach to introduce redundancy by allowing full pose read-out at all body joint locations breaks down for interacting and overlapping people, leading to supervision and \emph{read-out} conflicts in all location-map channels. 
Our selective introduction of redundancy restricts these conflicts to pose-map channels of similar limbs, \ie, wrist of one person in the proximity of a knee of another person cannot cause read-out conflicts because their pose is encoded in their respective pose-maps. 
If the complete pose was encoded at each joint location, there would be conflicts for each pair of proximate joints across people.
We now formally define \newLocationMaps (Sec.~\ref{sec:ORPM_formulation}) and explain the inference process (Sec.~\ref{sec:pose_inference}). 

%
\subsection{Formulation}
\label{sec:ORPM_formulation}
\begin{figure*}
  \begin{center}
  \includegraphics[width=0.98\linewidth, trim=0 0 0 10, clip]{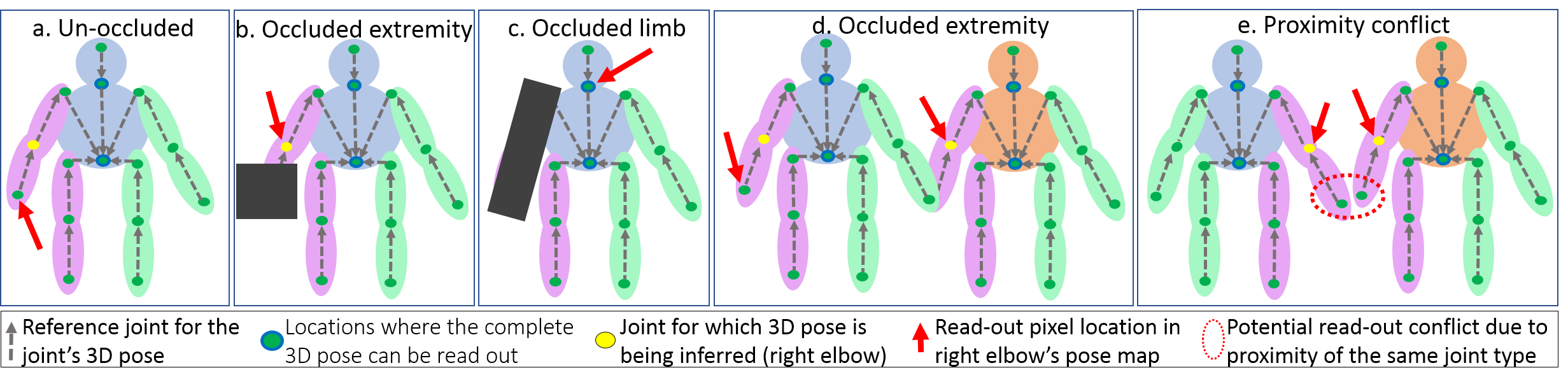}
  \end{center}
  \vspace{-0.4cm}
  \caption
  {
  Example of the choice of read-out pixel location for right elbow pose under various scenarios. First the complete body pose is read out at one of the torso locations. a.)~If the limb extremity is un-occluded, the pose for the entire limb is read out at the extremity (wrist), b.)~If the limb extremity is occluded, the pose for the limb is read out at the joint location further up in the joint hierarchy (elbow), c.)~If the entire limb is occluded, we retain the base pose read out at one of the torso locations (neck), d.)~Read-out locations indicated for inter-person interaction, e.)~If two joints of the same type (right wrist here) overlap or are in close proximity, limb pose read-out is done at a safer isolated joint further up in the hierarchy.
  }  
  \vspace{-0.3cm}
  \label{fig:pose_representation}
\end{figure*}

Given a monocular RGB image $\mathcal{I}$ of size $W \times H$, we seek to estimate the 3D pose $\mathcal{P} = \{\mathbf{P}_i\}_{i=1}^m$ for each of the $m$ persons in the image. 
Here, $\mathbf{P}_i \in \mathbb{R}^{3 \times n}$ describes the 3D locations of the $n = 17$ body joints of person $i$. The body joint locations are expressed relative to the parent joints as indicated in Fig.~\ref{fig:pose_representation} and converted to pelvis-relative locations for evaluation. 
We first decompose the body into pelvis, neck, head, and a set of limbs: 
$ 
\quad L = \{ \{\text{shoulder}_s, \text{elbow}_s, \text{wrist}_s\}, \{\text{hip}_s, \text{knee}_s, \text{ankle}_s\}\; |\; s \in \{\text{right}, \text{left}\}\} \:
$.
The 3D locations of the joints are then encoded in the occlusion-robust pose-maps denoted by $\mathcal{M} = \{ \mathbf{M}_j \}_{j=1}^n$, where $\mathbf{M}_j \in \mathbb{R}^{W \times H \times 3}$.
In contrast to simple location-maps, the ORPM $\mathbf{M}_j$ stores the 3D location of joint $j$ not only at this joint's 2D pixel location $(u,v)_j$ but at a set of 2D locations 
$
\rho(j) = \{(u,v)_{\text{neck}}, (u,v)_{\text{pelvis}}\} \cup \{(u,v)_k\}_{k \in {limb(j)}}
$,
where:  
\begin{equation}
limb(j) = 
\begin{cases}
l, &\text{if} \;\exists l \in L \;\text{with}\; j \in l \\
\{\text{head}\}, & \text{if} \; j = \text{head} \\
\emptyset, &\text{otherwise} 
\end{cases} \quad .
\end{equation}
Note that---since joint $j$ of all persons $i$ is encoded in 
$\mathbf{M}_j$---it can happen that read-out locations coincide for different people, \ie, $\rho_{i1}(j) \cap \rho_{i2}(j) \neq \emptyset$.
In this case, $\mathbf{M}_j$ contains information about the person closer to the camera at the overlapping locations.
However, due to our built-in redundancy in the ORPMs, a pose estimate for the partially occluded person can still be obtained at other available read-out locations. 

To estimate where the pose-maps can be read out, we make use of 2D joint \emph{heatmaps} $\mathcal{H} = \{ \mathbf{H}_j \in \mathbb{R}^{W \times H} \}_{j=1}^n$ predicted by our network.
Additionally, we estimate \emph{part affinity fields} $\mathcal{A} = \{ \mathbf{A}_j \in \mathbb{R}^{W \times H \times 2} \}_{j=1}^{n}$ which represent a 2D vector field pointing from a joint of type $j$ to its parent~\cite{cao_affinity_2017}. 
This facilitates association of 2D detections in the heatmaps (and hence read-out locations for the ORPMs) to person identities and enables per-person read-outs when multiple people are present.
Note that we predict a fixed number of maps ($n$ heatmaps, $3n$ pose-maps, and $2n$ part affinity fields) \change{in a single forward pass} irrespective of the number of persons in the scene, jointly encoding the 2D and 3D pose for all subjects, \ie, our network is \emph{single-shot}.

\subsection{Pose Inference}
\label{sec:pose_inference}
Read-out of 3D pose of multiple people from \newLocationMaps starts with inference of 2D joint locations $\mathcal{P}^{2D} = \{\mathbf{P^{2D}}_i\}_{i=1}^m$ with $\mathbf{P^{2D}}_i = \{(u,v)^i_j\}_{j=1}^n$ and joint detection confidences $\mathcal{C}^{2D} = \{\mathbf{C^{2D}}_i \in \mathbb{R}^{n}\}_{i=1}^m$ for each person $i$ in the image. 
Explicit 2D joint-to-person association is done with the predicted heatmaps $\mathcal{H}$ and part affinity fields $\mathcal{A}$ using the approach of Cao \etal~\cite{cao_affinity_2017}.
Next, we use the 2D joint locations $\mathcal{P}^{2D}$ and the joint detection confidences $\mathcal{C}^{2D}$ in conjunction with ORPMs $\mathcal{M}$ to infer the 3D pose of all persons in the scene.

\parahead{Read-Out Process} 
By virtue of the \newLocationMaps we can read out 3D joint locations at select multiple pixel locations as described above. 
We define \emph{extremity joints}: the wrists, the ankles, and the head. 
The neck and pelvis 2D detections are usually reliable, these joints are most often not occluded and lie in the middle of the body. 
Therefore, we start reading the full base pose at the neck location. 
If the neck is \emph{invalid} (as defined below) then the full pose is read at the pelvis instead. 
If both of these joints are invalid, we consider this person as not visible in the scene and we do not predict the person's pose. 
While robust, full poses read at the pelvis and neck tend to be closer to the average pose in the training data. 
Hence, for each limb, we continue by reading out the limb pose at the extremity joint. 
Note again that the \emph{complete} \emph{limb} pose can be accessed at any of that limb's 2D joint locations.
If the extremity joint is valid, the limb pose replaces the corresponding elements of the base pose. 
If the extremity joint is invalid however, we walk up the kinematic chain and check the other joints of this limb for validity. 
If all joints of the limb are invalid, the base pose cannot be further refined.
This read-out procedure is illustrated in Fig.~\ref{fig:pose_representation} and algorithmically described in the supplementary document.
 
\parahead{2D Joint Validation}
We declare a 2D joint location $\mathbf{P^{2D}}_{i}^j = (u,v)^i_j$ of person $i$ as \emph{valid read-out location} iff (1) it is un-occluded, \ie, has confidence value higher than a threshold $t_C$, and (2) it is sufficiently far ($\geq t_D$) away from all read-out locations of joint $j$ of other individuals:
\begin{align}
valid(\mathbf{P^{2D}}_{i}^j) \  \Leftrightarrow \  
& \mathbf{C^{2D}}_{i}^j > t_C \ \wedge \nonumber\ || a - \mathbf{P^{2D}}_{i}^j||_2 \geq t_D \ \\
& \forall \bar{i}=[1\!:\!m], \: \bar{i} \neq i. \ \forall a \in \rho_{\bar{i}}(j).  
\end{align}

Our \newLocationMap formulation together with the occlusion-aware inference strategy with limb refinement enables us to obtain accurate poses even for strongly occluded body parts while exploiting all available information if individual limbs are visible.
We validate our performance on occluded joints and the importance of limb refinement on our new test set (see Sec.~\ref{sec:results}).

\subsection{Network and Training Details}
Our network has ResNet-50~\cite{he_resnet_cvpr2016} as the core, after which we split it into two---a \emph{2DPose+Affinity} stream and a \emph{3DPose} stream. 
The core network and the first branch are trained on MS-COCO~\cite{lin_coco_eccv14} and the second branch is trained with MPI-INF-3DHP or MuCo-3DHP as per the scenario.
Training and architectural specifics are in the supplementary document. 

The \emph{2DPose+Affinity} stream predicts the 2D heatmaps $\mathcal{H}_{COCO}$ for the MS-COCO body joint set, and part affinity fields $\mathcal{A}_{COCO}$.
The \emph{3DPose} stream predicts 3D \newLocationMaps $\mathcal{M}_{MPI}$ as well as 2D heatmaps $\mathcal{H}_{MPI}$ for the MPI-INF-3DHP~\cite{mehta_mono_3dv17} joint set, which has some overlap with the MS-COCO joint set. 
For pose read-out locations as described previously, we restrict ourselves to the common minimum joint set between the two, indicated by the circles in Fig.~\ref{fig:pose_representation}. 

\parahead{Loss}
The 2D heatmaps $\mathcal{H}_{COCO}$ and $\mathcal{H}_{MPI}$ are trained with per-pixel $L2$ loss comparing the predictions to the reference which has unit peak Gaussians with a limited support at the ground truth 2D joint locations, as is common. The part affinity fields $\mathcal{A}_{COCO}$ are similarly trained with a per-pixel $L2$ loss, using the framework made available by Cao \etal~\cite{cao_affinity_2017}.
While training \newLocationMaps with our \emph{MuCo-3DHP}, per joint type $j$, for all subjects $i$ in the scene, a per-pixel $L2$ loss is enforced in the neighborhood of all possible read-out locations $\rho_i(j)$. 
The loss is weighted by a limited support Gaussian centered at the read-out location.

\begin{table*}[t]
\renewcommand{\tabcolsep}{1.5pt}
\centering
  \caption
  { Sequence-wise evaluation of our method and LCR-net\cite{rogez_lcr_cvpr17} on multi-person 3D pose test set \emph{MuPoTS-3D}. We report both (a) the overall accuracy (3DPCK), and (b) accuracy only for person annotations matched to a prediction 
  }  
\vspace{0.2cm}
\label{tbl:test_set_evaluation}
\resizebox{1.02\linewidth}{!}{
\begin{tabular}{c|c||c|c|c|c|c|c|c|c|c|c|c|c|c|c|c|c|c|c|c|c||c}
                                          &         & TS1  & TS2  & TS3  & TS4  & TS5  & TS6  & TS7  & TS8  & TS9  & TS10 & TS11 & TS12 & TS13 & TS14 & TS15 & TS16 & TS17 & TS18 & TS19 & TS20 & \textbf{Total} \\ \hline
\multicolumn{1}{l|}{\multirow{2}{*}{a.)}} & LCR-net\T & 67.7 & 49.8 & 53.4 & 59.1 & 67.5 & 22.8 & 43.7 & 49.9 & 31.1 & 78.1 & 50.2 & 51.0 & 51.6 & 49.3 & 56.2 & 66.5 & 65.2 & 62.9 & 66.1 & 59.1 & 53.8  \\
\multicolumn{1}{l|}{}                 & Ours    & \textbf{\textbf{81.0}} & \textbf{59.9} & \textbf{64.4} & \textbf{62.8} & \textbf{68.0} & \textbf{30.3} & \textbf{65.0} & \textbf{59.2} & \textbf{64.1} & \textbf{83.9} & \textbf{67.2} & \textbf{68.3} & \textbf{60.6}     & \textbf{56.5} & \textbf{69.9} & \textbf{79.4} & \textbf{79.6} & \textbf{66.1} & \textbf{66.3} & \textbf{63.5} & \textbf{65.0}  \\ \hline
\multicolumn{1}{l|}{\multirow{2}{*}{b.)}} &LCR-net \T& 69.1 & \textbf{67.3} & 54.6 & 61.7 & \textbf{74.5} & 25.2 & 48.4 & \textbf{63.3} & \textbf{69.0} & 78.1 & 53.8 & 52.2 & 60.5 & 60.9 & 59.1 & 70.5 & 76.0 & 70.0 & \textbf{77.1} & 81.4 & 62.4\\ 
\multicolumn{1}{l|}{}   & Ours    & \textbf{81.0} & 64.3 & \textbf{64.6} & \textbf{63.7} &     {73.8} & \textbf{30.3} & \textbf{65.1} & 60.7 & 64.1 & \textbf{83.9} & \textbf{71.5} & \textbf{69.6} & \textbf{69.0} & \textbf{69.6} & \textbf{71.1} & \textbf{82.9} & \textbf{79.6} & \textbf{72.2} & {76.2} & \textbf{85.9} & \textbf{69.8}
\end{tabular}
}
\vspace{-0.3cm}
\end{table*}

\section{Results and Discussion}
\label{sec:results}
The main goal of our method is \emph{multi-person} 3D pose estimation in general scenes, which exhibits specific and more difficult challenges than single-person pose estimation.
However, we validate the usefulness of our \newLocationMap formulation on the single-person pose estimation task as well. To validate our approach, we perform extensive experiments on our proposed multi-person test set \emph{MuPoTS-3D}, as well as two publicly available single-person benchmarks, namely \emph{Human3.6m}~\cite{ionescu_human36_pami14} and \emph{MPI-INF-3DHP}~\cite{mehta_mono_3dv17}.
Fig.~\ref{fig:teaser} presents qualitative results of our method showcasing the ability to handle complex in-the-wild scenes with strong inter-person occlusion.
An extensive collection of qualitative results is provided in the supplementary document and video.

\subsection{Comparison with Prior Art} 
\label{sec:comparisons}
For sequences with strong occlusion, we obtain much better results than the state-of-the-art. For un-occluded sequences our results are comparable to methods designed for single person. 
We outperform the only other multi-person method (LCR-net~\cite{rogez_lcr_cvpr17}) quantitatively and qualitatively on both single-person and multi-person tasks.
For fairness of comparison, in all evaluations, we re-target the predictions from LCR-net~\cite{rogez_lcr_cvpr17} to a skeleton with bone-lengths matching the ground truth.

\parahead{Multi-Person Pose Performance}
We use our proposed \emph{MuPoTS-3D} (see Sec.~\ref{sec:test_set}) to evaluate multi-person 3D pose performance in general scenes for our approach and LCR-net~\cite{rogez_lcr_cvpr17}.
In addition, we evaluate VNect~\cite{VNect_SIGGRAPH2017} on images cropped with the ground truth bounding box around the subject. 
We evaluate for all subjects that have 3D pose annotations available. If an annotated subject is missed by our method, or by LCR-net, we consider all of its joints to be incorrect in the 3DPCK metric.
Table~\ref{tbl:test_set_evaluation}(a) reports the 3DPCK metric for all 20 sequences when taking all available annotations into account. 
Our method performs significantly better than LCR-net for most sequences, while being comparable for a few, yielding an overall improved performance of 65.0 3DPCK vs 53.8 3DPCK for LCR-net. 
We provide a joint-wise breakdown of the overall accuracy in the supplementary document.

Overall, our approach detects 93\% of the annotated subjects, whereas LCR-net was successful for 86\%. This is an additional indicator of performance. 
Even ignoring the undetected annotated subjects, our approach outperforms LCR-net in terms of 3D pose error (69.8 vs 62.4 3DPCK, and 132.5 vs 146 mm MPJPE).

VNect is evaluated on ground truth crops of the subjects, and therefore it operates at a 100\% detection rate. In contrast, we do not use ground truth crops, and missed detections by our method count as all joints wrong. Despite this our method achieves
better accuracy (65.0 vs 61.1 3DPCK, 30.1 vs 27.6 AUC).

\parahead{Single-Person Pose Performance}
On the MPI-INF-3DHP dataset (see Table~\ref{tbl:our_testset}) we compare our method trained on MPI-INF-3DHP (single-person) and MuCo-3DHP (multi-person) to three single-person methods---VNect, Zhou \etal~\cite{zhou2017towards}, Mehta \etal~\cite{mehta_mono_3dv17}---and LCR-net as the only other multi-person approach.
Our method trained on multi-person data (73.4 3DPCK) performs marginally worse than our single-person version (75.2 3DPCK) due to the effective loss in network capacity when training on harder data.
Nevertheless, both our versions consistently outperform Zhou \etal~(69.2 3DPCK) and LCR-net (59.7 3DPCK) over all metrics. 
LCR-net predictions have a tendency to be conservative about the extent of articulation of limbs as shown in Fig.~\ref{fig:qualitative_results}.
In comparison to VNect (76.6 3DPCK) and Mehta \etal (75.7 3DPCK), our method (75.2 3DPCK) achieves an average accuracy that is on par. 
Our approach outperforms existing methods by $\approx$3-4 3DPCK for activities which exhibit significant self- and object-occlusion like \emph{Sit on Chair} and \emph{Crouch/Reach}. For the full activity-wise breakdown, see the supplemental document.

For detailed comparisons on Human3.6m~\cite{ionescu_human36_pami14}, refer to the supplementary document.
Our approach at 69.6mm MPJPE performs $\approx$17mm better than LCR-net~\cite{rogez_lcr_cvpr17} (87.7mm), and outperforms the VNect location-map~\cite{VNect_SIGGRAPH2017} (80.5mm) formulation by $\approx$10mm. Our results are comparable to the recent state-of-the-art results of Pavlakos \etal ~\cite{pavlakos_volumetric_cvpr17} (67.1mm), Martinez \etal \cite{martinez20173dbaseline} (62.9mm), Zhou \etal~\cite{zhou2017towards} (64.9mm), Mehta \etal~\cite{mehta_mono_3dv17} (68.6mm) and Tekin \etal~\cite{tekin_fusion_arxiv16} (70.81mm), and better than the recent results from Nie \etal~\cite{nie2017monocular} (79.5mm) and Tome \etal~\cite{tome_lifting_2017} (88.39mm). 
\begin{figure}
\centering
  \includegraphics[width=0.95\linewidth]{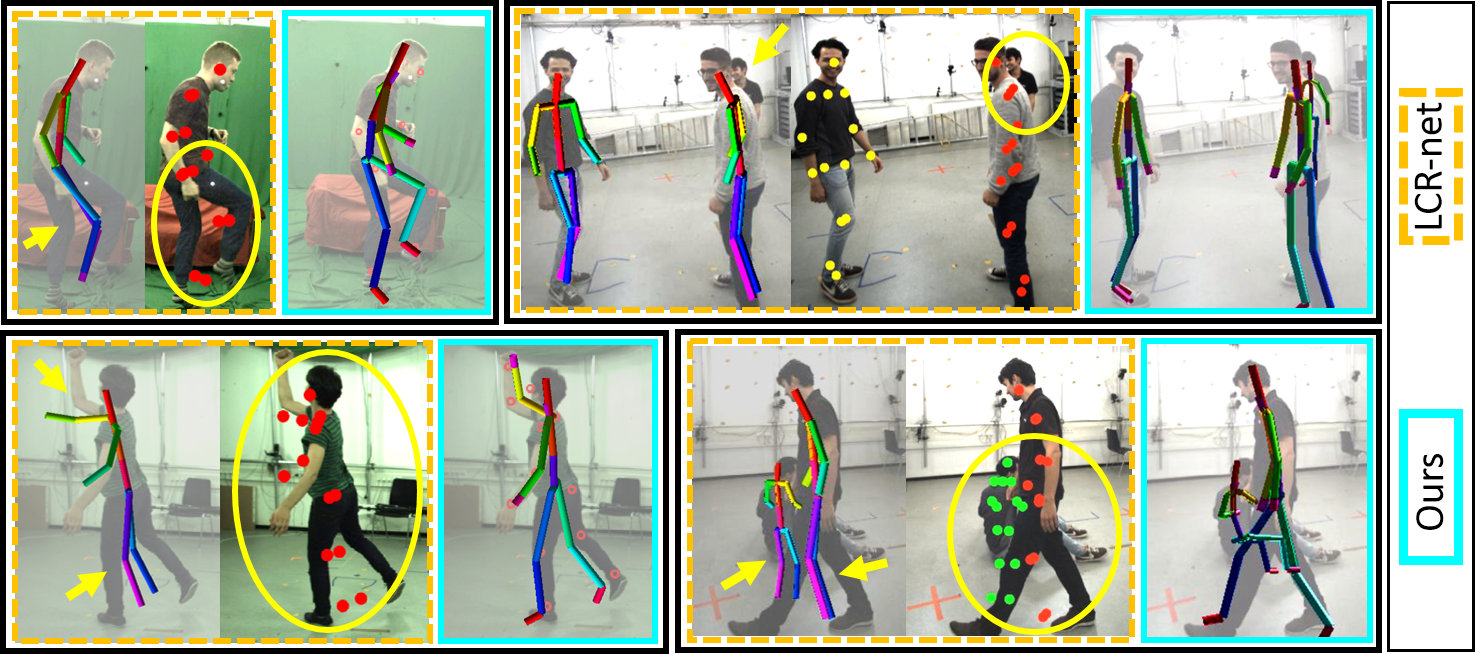}
  \vspace{-0.0cm}
  \caption
  {Qualitative comparison of LCR-net~\cite{rogez_lcr_cvpr17} and our method. LCR-net output is limited in the extent of articulation of limbs, tending towards neutral poses. LCR-net also has more detection failures under significant occlusion. 
  }  
  \label{fig:qualitative_results}
  \vspace{-0.5cm}
\end{figure}

\textbf{Occlusion evaluation: }
To demonstrate the occlusion robustness of our method, we create synthetic random occlusions on the MPI-INF-3DHP test set. 
The synthetic occlusions cover about 14\% of the joints. 
Our single-person and multi-person variants both outperform VNect for occluded joints (62.8 vs 64.0 vs 53.2 3DPCK) by a large margin, and are comparable for un-occluded joints (67.0 vs 71.0 vs 69.4 3DPCK). Note again that the single-person variant is trained on similar data as VNect and the occlusion robustness is inherent to the formulation.
See the supplemental document for a more detailed breakdown by test sequence.

We use the per-joint occlusion annotations from \emph{MuPoTS-3D} to further assess LCR-net and our approach under occlusion. 
Considering both self- and inter-personal occlusions, $\approx$23.7 \% of the joints of all subjects are occluded.
Our method is more robust than LCR-net on both occluded (48.7 vs 42 3DPCK) and un-occluded (70.0 vs 57.5 3DPCK) joints. 

\subsection{Ablative Analysis}
\label{sec:ablative_analysis}
We validate the improvement provided by our limb refinement strategy on MPI-INF-3DHP test set.
We empirically found that the base pose read out at one of the torso joints tends towards the mean pose of the training data.
Hence, for poses with significant articulation of the limbs, the base pose does not provide an accurate pose estimate, especially for the end effectors.
On the other side, the limb poses read out further down in the kinematic chain, \ie, closer to the extremities, include more detailed articulation information for that limb.
Our proposed read-out process exploits this fact, significantly improving overall pose quality by limb refinement when limbs are available.
Table~\ref{tbl:our_testset} shows that the benefit of the full read-out is consistent over all metrics and valid for our method independent of whether it is trained on single-person (MPI-INF-3DHP) or multi-person data (MuCu-3DHP), with a $\approx$10 3DPCK advantage over torso read-out. See Fig.~2,3 in the supplementary document.

\begin{table}[t]
\centering
\renewcommand{\tabcolsep}{3.5pt}

\centering
\caption{Comparison of results on MPI-INF-3DHP~\cite{mehta_mono_3dv17} test set. We report the \emph{Percentage of Correct Keypoints measure in 3D} (@150mm) for select activities, and the total 3DPCK and the Area Under the Curve for \underline{all} activities. Complete activity-wise breakdown in the supplementary document
}
\vspace{-0.0cm}
\label{tbl:our_testset}
\resizebox{0.95\linewidth}{!}{
\begin{tabular}{l||cc||cc}
            & \multicolumn{1}{c|}{Sit} & \multicolumn{1}{c||}{Crouch} & \multicolumn{2}{c}{Total}        \\ \cline{2-5}
 Method \T &   \multicolumn{1}{c|}{PCK} &   \multicolumn{1}{c||}{PCK}&   \multicolumn{1}{c|}{PCK}  & \multicolumn{1}{c}{AUC}         \\ \hline \hline
\multirow{1}{*}{\begin{tabular}[c]{@{}c@{}}VNect~\cite{VNect_SIGGRAPH2017}\end{tabular}}    \T &  \multicolumn{1}{c|}{{74.7}} &  {72.9} &  \multicolumn{1}{c|}{\textbf{76.6}} & \multicolumn{1}{c}{\textbf{40.4}}                 \\ \hline
\multirow{1}{*}{\begin{tabular}[c]{@{}c@{}}LCR-net~\cite{rogez_lcr_cvpr17}\end{tabular}}    \T  &  \multicolumn{1}{c|}{{58.5}} &  {69.4} & \multicolumn{1}{c|}{59.7} & \multicolumn{1}{c}{27.6}                 \\\hline
\multirow{1}{*}{\begin{tabular}[c]{@{}c@{}}Zhou et al.\cite{zhou2017towards}\end{tabular}}    \T &  \multicolumn{1}{c|}{{60.7}} &  {71.4} & \multicolumn{1}{c|}{69.2} & \multicolumn{1}{c}{32.5}                 \\\hline
\multirow{1}{*}{\begin{tabular}[c]{@{}c@{}}Mehta et al.\cite{mehta_mono_3dv17}\end{tabular}}   \T &  \multicolumn{1}{c|}{{74.8}} &  {73.7} &  \multicolumn{1}{c|}{75.7} & \multicolumn{1}{c}{39.3}                 \\\hline \hline
\multirow{1}{*}{\begin{tabular}[c]{@{}c@{}}Our Single-Person (Torso)\end{tabular}}    \T &  \multicolumn{1}{c|}{{69.1}} &  {68.7} &  \multicolumn{1}{c|}{{65.6}} & \multicolumn{1}{c}{{32.6}}                 \\ \cline{2-5}
\multirow{1}{*}{\begin{tabular}[c]{@{}c@{}}Our Single-Person (Full)\end{tabular}}\T                                                                                         &  \multicolumn{1}{c|}{\textbf{77.8}} &  \textbf{77.5} & \multicolumn{1}{c|}{75.2} & \multicolumn{1}{c}{37.8}                 \\ \hline 
\multirow{1}{*}{\begin{tabular}[c]{@{}c@{}}Our Multi-Person (Torso)\end{tabular}}    \T &  \multicolumn{1}{c|}{{64.6}} &  {65.8}&  \multicolumn{1}{c|}{{63.6}} & \multicolumn{1}{c}{{31.1}}                 \\ \cline{2-5}
\multirow{1}{*}{\begin{tabular}[c]{@{}c@{}}Our Multi-Person (Full)\end{tabular}}\T        &  \multicolumn{1}{c|}{{75.9}} &  {73.9} & \multicolumn{1}{c|}{73.4} & \multicolumn{1}{c}{36.2}                 \\ \hline 
\end{tabular}
}
\vspace{-0.4cm}
\end{table}

\section{Limitations and Future Work}
As discussed in Section~\ref{sec:method}, we handle overlapping joints of the same type by only supervising the one closest to the camera. However, when joints of the same type are in close proximity (but not overlapping) the ground-truth ORPM for those may transition sharply from one person to the other, which are hard to regress and may lead to inaccurate predictions.
One possible way to alleviate the issue is to increase the resolution of the output maps. Another source of failures is when 2D joints are mis-predicted or mis-associated.
Furthermore, we have shown accurate root-relative 3D pose estimation, but estimating the relative sizes of people is challenging and remains an open problem for future work.
While the compositing based \emph{MuCo-3DHP} covers many plausible scenarios, further investigation into capturing/generating true person-person interactions at scale would be an important next step.
\section{Conclusion}
Multi-person 3D pose estimation from monocular RGB is a challenging problem which has not been fully addressed by previous work.
Experiments on single-person and multi-person benchmarks show that our method, relying on a novel occlusion robust pose formulation (ORPM), works well to estimate the 3D pose even under strong inter-person occlusions and human--human interactions better than previous approaches.
The method has been trained on our new multi-person dataset (\emph{MuCo-3DHP}) synthesized at scale from existing single-person images with 3D pose annotations. Our method trained on this dataset generalizes well to real world scenes shown in our \emph{MuPOTS-3D} evaluation set.
We hope further investigation into monocular multi-person pose estimation would be spurred by the proposed training and evaluation data sets.

\clearpage
\begin{center}
\textbf{\large Supplementary Document:\\Single-Shot Multi-Person 3D Pose \\Estimation From Monocular RGB}
\end{center}
\setcounter{equation}{0}
\setcounter{figure}{0}
\setcounter{table}{0}
\setcounter{section}{0}
\section{Read-out Process}
An algorithmic description of the read-out process is provided in Alg.~\ref{algo:readout}.
\begin{algorithm}
\caption{3D Pose Inference} \label{algo:readout}
\begin{algorithmic}[1]
\State{Given: $\mathcal{P}^{2D}$, $\mathcal{C}^{2D}$, $\mathcal{M}$}
\ForAll{$i \in (1..m)$}
\If{ $\mathbf{C}^{2D}_i[k]> thresh$,  $k \in \{pelvis, neck\}$}
\State Person $i$ is detected 
\ForAll{joints $j \in (1..n)$}
\State $rloc = \mathbf{P}^{2D}_i[k]$
\State $\mathbf{P}_i[:,j]$ = \Call{ReadLocMap}{j, rloc}
\EndFor
\ForAll{limbs $l \in \{arm_l, arm_r, leg_l, leg_r, head\}$}
\For{$j = \Call{getExtremity}{l}; j \notin \{pelvis, neck\}; j = parent(j)$} 
\If{\Call{isValidReadoutLoc}{i, j}} 
\State \Call{refineLimb}{l, $\mathbf{P}^{2D}_i[j]$}
\State \textbf{break}
\EndIf
\EndFor
\EndFor
\Else
\State No person detected
\EndIf
\EndFor

\Function{getExtremity}{limb l}
\If{$l = leg_s$}
\Return $ankle_s$
\Else{}
\If{$l = arm_s$}
\Return $wrist_s$
\Else{}
\Return $head$
\EndIf
\EndIf
\EndFunction

\Function{ReadLocMap}{joint j, 2DLocation rloc} 
\State $rloc = rloc / locMap\_scale\_factor$
\State \Return $\mathbf{M}_j[rloc]$
\EndFunction

\Function{refineLimb}{limb l, 2DLocation rloc}
\ForAll{joints $b \in$ limb $l$}
\State $\mathbf{P}_i[:,b]$=\Call{ReadLocMap}{b, rloc} 
\EndFor
\EndFunction

\Function{isValidReadoutLoc}{person i, joint j} 
\If{ $(\mathbf{C}^{2D}_i[j]> 0)$}
\State \Return \Call{isIsolated}{i,j} 
\Else
\State \Return 0
\EndIf
\EndFunction

\Function{isIsolated}{person i, joint j} 
\State{$isol = 1$}
\ForAll{persons $\bar{i} \in (1..m), \bar{i} \neq i$}
\ForAll{2DLocations $a \in \rho_{\bar{i}}(j)$}
\If{$||a - \mathbf{P}^{2D}_{i}[j]||_2 < isoThresh$ }
\State{$isol = 0$}
\State \textbf{break}
\EndIf
\EndFor
\EndFor
\State \Return $isol$
\EndFunction
\end{algorithmic}
\end{algorithm}
\section{Network Details}

\subsection{Architecture}
A visualization if our network architecture using the web-based visualization tool \emph{Netscope} can be found at: \url{http://ethereon.github.io/netscope/#/gist/069a592125c78fbdd6eb11fd45306fa0}.

\begin{figure}
  \centering
  \includegraphics[width=1.0\linewidth]{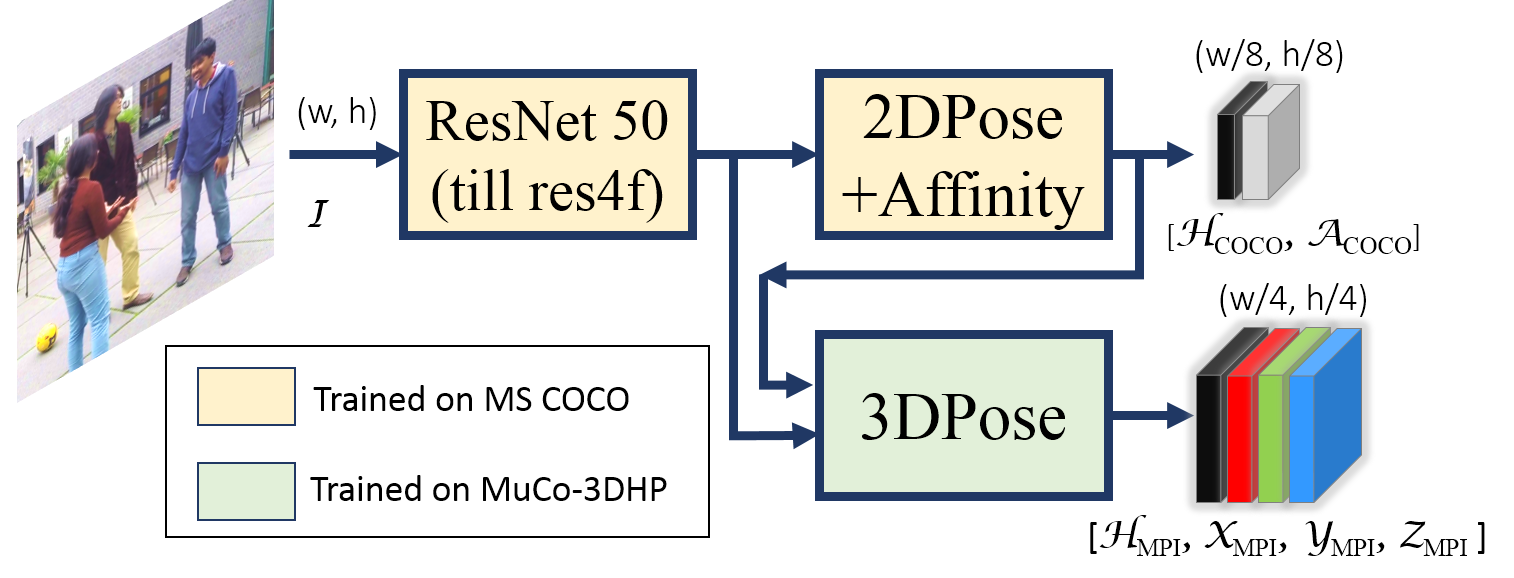}
  \vspace{-0.2cm}
  \captionof{figure}
  {
  The network architecture with \emph{2DPose+Affinity} branch predicting the 2D \emph{heatmaps} $\mathcal{H}_{COCO}$ and \emph{part affinity maps} $\mathcal{A}_{COCO}$ with a spatial resolution of $(W/8,H/8)$, and \emph{3DPose} branch predicting 2D \emph{heatmaps} $\mathcal{H}_{MPI}$ and \newLocationMaps $\mathcal{M}_{MPI}$ with a spatial resolution of $(W/4,H/4)$, for an input image with resolution $(W,H)$.
  }  
  \vspace{-0.0cm}
  \label{fig:net_arch_branch}
\end{figure}

\subsection{Data}
We use 12 out of the 14 available camera viewpoints (using only 1 of the 3 available top down views) in MPI-INF-3DHP~\cite{mehta_mono_3dv17} training set, and create 400k composite frames of MuCo-3DHP, of which half are without appearance augmentation. 
For training, we crop around the subject closest to the camera, and apply rotation, scale, and bounding-box jitter augmentation.
Since the data was originally captured in a relatively restricted space, the likelihood of there being multiple people visible in the crop around the main person is high. The combination of scale augmentation, bounding-box jitter, and cropping around the subject closest to the camera results in many examples with truncation from the frame boundary, in addition to the inter-person occlusions occurring naturally due to the compositing.
\subsection{Training}
We train our network using the Caffe~\cite{jia_caffe_ICM} framework.
The core network's weights were initialized with those trained for 2D body pose estimation on MPI~\cite{andriluka_mpii2d_cvpr14} and LSP~\cite{johnson_lsp_bmvc10,johnson_lspet_cvpr11} datasets as done in~\cite{mehta_mono_3dv17}.
The core network and the \emph{2DPose + Affinity} branch are trained for multi-person 2D pose estimation using the framework provided by Cao et al.~\cite{cao_affinity_2017}. We use the AdaDelta solver, with a momentum of 0.9 and weight decay multiplier of 0.005, and a batch size of 8. We train for 640k iterations with a cyclical learning rate ranging from 0.1 to 0.000005.
The \emph{3DPose} branch is trained with the core network and \emph{2DPose + Affinity} branch weights frozen. We use a batch size of 6 and train for 360k iterations with a cyclical learning rate ranging from 0.1 to 0.000001.
We empirically found that training the part affinity fields and occlusion-robust pose-maps at lower resolution (see Fig.~\ref{fig:net_arch_branch}) leads to better results.

\section{Joint-wise Analysis}
\begin{figure}
\centering
  \includegraphics[width=1.0\linewidth, trim={1.3cm 0 0 0}, clip]{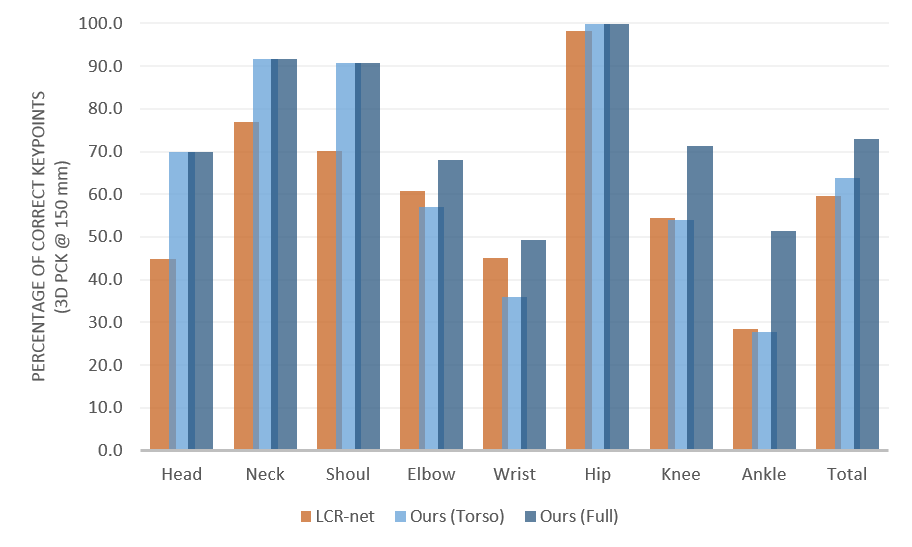}
  \vspace{-0.4cm}
  \caption
  {Joint-wise accuracy comparison of our method and LCR-net~\cite{rogez_lcr_cvpr17} on the single person MPI-INF-3DHP test set. 3D Percentage of Correct Keypoints (@150mm) as the vertical axis. LCR-net predictions were mapped to the ground truth bone lengths for fairness of comparison.}
  \vspace{-0.3cm}
  \label{fig:jointwise_single}
\end{figure}

\begin{figure}
\centering
  \includegraphics[width=0.90\linewidth]{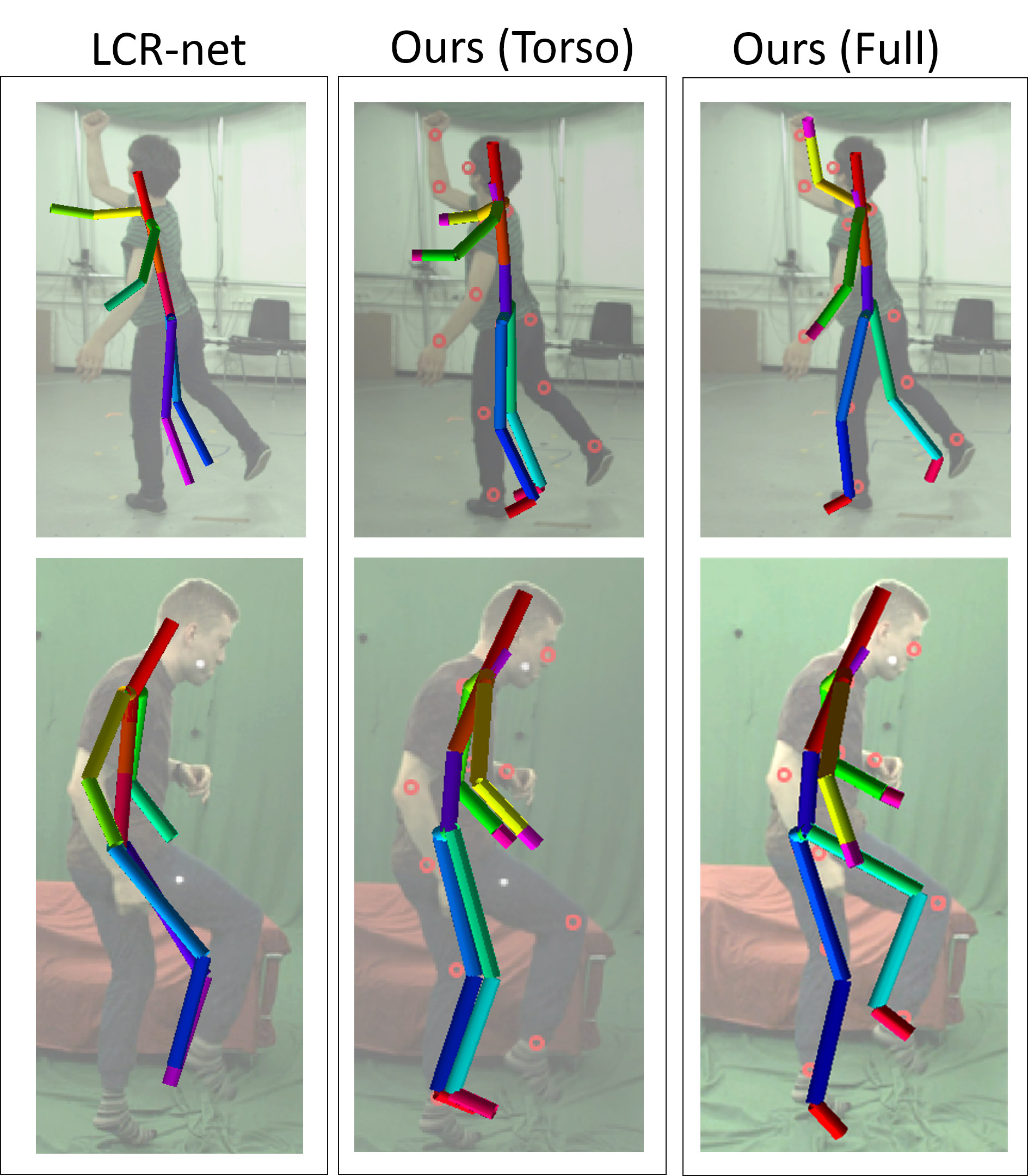}
  \vspace{-0.0cm}
  \caption
  {Qualitative comparison of LCR-net~\cite{rogez_lcr_cvpr17} and our method. LCR-net predictions are limited in terms of the extent of articulation of limbs, tending towards neutral poses. For our method, the base pose read out at the torso is similarly limited in terms of degree of articulation of limbs, and our full read-out addresses the issue. 
  }  
  \label{fig:lcr_torso_full}
  \vspace{-0.0cm}
\end{figure}

Figure \ref{fig:jointwise_single} shows joint-wise accuracy comparison of our approach with LCR-net~\cite{rogez_lcr_cvpr17} on the single person MPI-INF-3DHP test set. For limb joints (elbow, wrist, knee, ankle) LCR-net performs comparably or better than our torso-only readout, but our full readout performs significantly better. See Figure~\ref{fig:lcr_torso_full}.

Figure \ref{fig:jointwise_multi} shows joint-wise accuracy comparison of our approach with LCR-net on our proposed multi-person 3D pose test set. We see that our approach obtains a better accuracy for all joint types for most sequences, only performing worse than LCR-net for a select few joint types on certain sequences (TestSeq18,19,20).


\begin{figure}
\centering
  \includegraphics[width=0.90\linewidth]{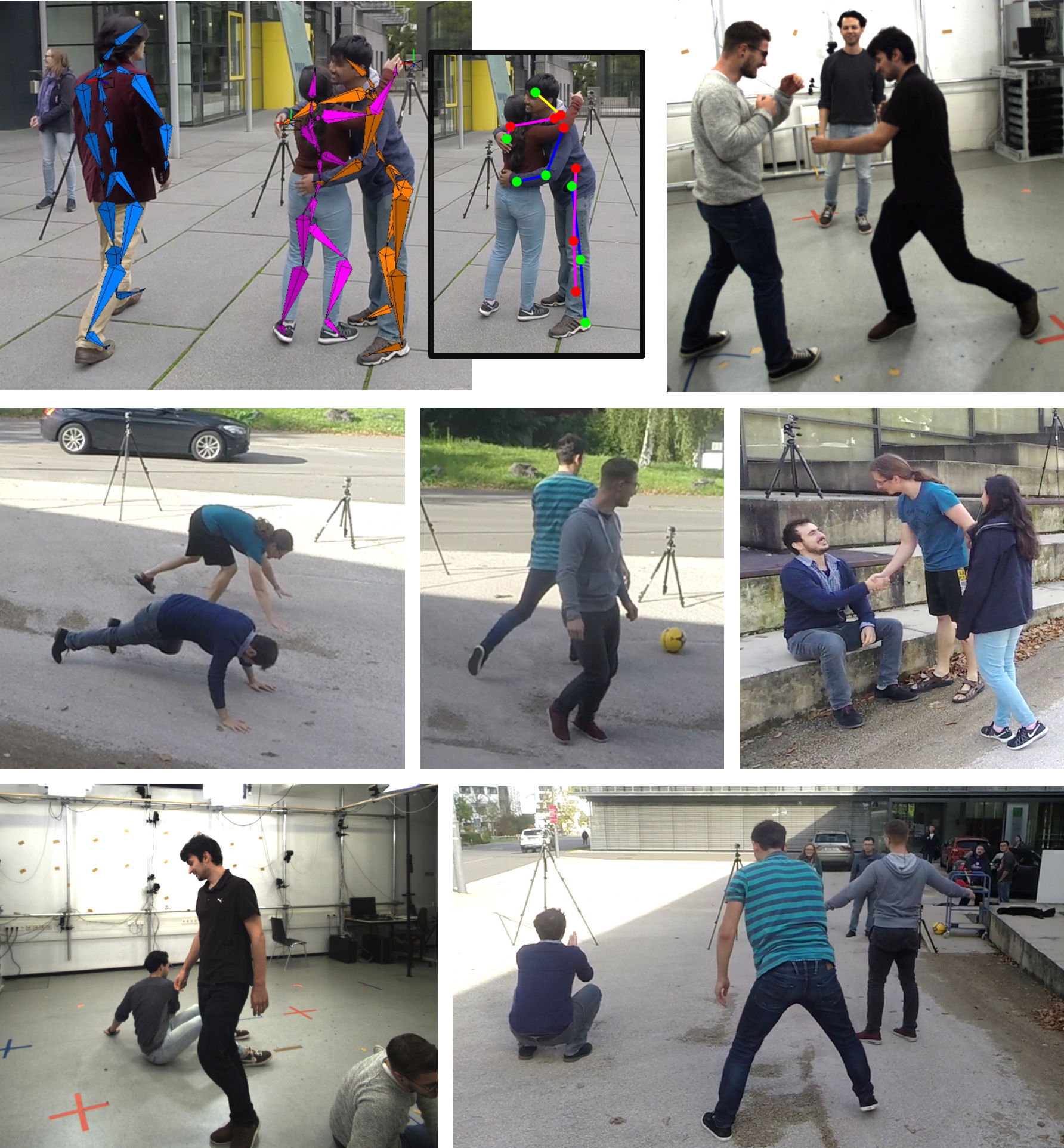}
  \vspace{-0.0cm}
  \caption
  {
  Examples from our MuPoTS-3D evaluation set. Ground truth 3D pose reference and joint occlusion annotations are available for up to 3 subjects in the scene (shown here for the frame on the top right). The set covers a variety of scene settings, activities and clothing.
  }  
  \label{fig:test_set}
  \vspace{-0.5cm}
\end{figure}

\begin{table*}
\centering
\caption{Comparison of results on Human3.6m \cite{ionescu_human36_pami14}, for single un-occluded person. Human3.6m, subjects 1,5,6,7,8 used for training. Subjects 9 and 11, all cameras used for testing. Mean Per Joint Postion Error reported in mm 
 }
\label{tbl:s9_11_compare_full}
\resizebox{0.9\linewidth}{!}{
\begin{tabular}{@{}lcccccccr@{}}
\multicolumn{1}{l||}{}                      & \textbf{Direct}   & \textbf{Disc.} & \textbf{Eat}   & \textbf{Greet} & \textbf{Phone} & \textbf{Pose}     & \textbf{Purch.}                         & \textbf{Sit.}     \\ \hline \hline

\multicolumn{1}{l||}{Pavlakos et al \cite{pavlakos_volumetric_cvpr17}\T}            & 60.9 &  67.1        & 61.8  &  62.8        &  67.5       & 58.8      &  64.4 & 79.8       \\
\multicolumn{1}{l||}{Mehta et al \cite{mehta_mono_3dv17} }                                                                & 52.5        & 63.8   & 55.4        & 62.3   & 71.8   & 52.6    & 72.2                          & 86.2   \\  
\multicolumn{1}{l||}{Tome et al \cite{tome_lifting_2017} }      & 65.0       & 73.5  & 76.8       & 86.4  & 86.3  & 69.0   & 74.8                         & 110.2   \\  
\multicolumn{1}{l||}{Chen et al \cite{chen_2d_match_cvpr17}}                              & 89.9            & 97.6 & 90.0        & 107.9  & 107.3       & 93.6        & 136.1                  & 133.1       \\
\multicolumn{1}{l||}{Moreno et al \cite{moreno_distance_matrix_cvpr17}}                              & 67.5            & 79.0  & 76.5        & 83.1  & 97.4       & 74.6        &  72.0        &  102.4       \\
\multicolumn{1}{l||}{Zhou et al \cite{zhou2017towards}}                              & 54.8            & 60.7  &  58.2       & 71.4  &  62.0       & 53.8        &  55.6        &  75.2       \\
\multicolumn{1}{l||}{Martinez et al \cite{martinez20173dbaseline}}     & 51.8           &  56.2  &  58.1       & 59.0  &  69.5       & 55.2        &  58.1        &  74.0       \\
\multicolumn{1}{l||}{Tekin et al \cite{tekin_fusion_arxiv16}}     & 53.9          &  62.2  &  61.5       & 66.2  &  80.1       & 64.6        &  83.2        &  70.9       \\
\multicolumn{1}{l||}{Nie et al \cite{nie2017monocular}}     & 62.8          &  69.2  &  79.6       & 78.8  &  80.8       & 72.5        &  73.9        &  96.1       \\
\multicolumn{1}{l||}{VNect \cite{VNect_SIGGRAPH2017}\T}                              & 62.6            & 78.1  & 63.4        & 72.5  & 88.3       & 63.1        & 74.8                              & 106.6       \\
\multicolumn{1}{l||}{LCR-net \cite{rogez_lcr_cvpr17}}                              & 76.2            & 80.2  &  75.8        & 83.3  &  92.2       & 79.9        & 71.7                  & 105.9       \\ \hline\hline
\multicolumn{1}{l||}{VNect (with our setup) \T}                                                                & 65.52        & 78.8  & 64.8        & 75.0   & 85.2   & 66.4    & 88.1                          & 110.2   \\  
\multicolumn{1}{l||}{Our Single-Person\T}                                                                & {58.2}        & {67.3}   & {61.2}    & {65.7}   & {75.82}   & {62.2}    & {64.6}                          & {82.0}   \\  \\

\multicolumn{1}{l||}{\T}                      & \textbf{Sit} & \textbf{}   & \textbf{} & \textbf{}  & \textbf{}  & \textbf{Walk} & \multicolumn{1}{c||}{\textbf{Walk}} & \textbf{} \\ 
\multicolumn{1}{l||}{}                      & \textbf{Down} & \textbf{Smk.}   & \textbf{Photo} & \textbf{Wait}  & \textbf{Walk}  & \textbf{Dog} & \multicolumn{1}{c||}{\textbf{Pair}} & \textbf{Avg.} \\ \hline \hline

\multicolumn{1}{l||}{Pavlakos et al \cite{pavlakos_volumetric_cvpr17}\T}                              &  92.9       & 67.0   & 72.3    &  70.0    & 54.0 & 71.0  &  \multicolumn{1}{c||}{57.6}         &  67.1     \\
\multicolumn{1}{l||}{Mehta et al \cite{mehta_mono_3dv17} }                                                                & 120.6    & 66.0   & 79.8    & 64.0   &  48.9   & 76.8    & \multicolumn{1}{c||}{53.7}     & {68.6}  \\ 
\multicolumn{1}{l||}{Tome et al \cite{tome_lifting_2017} }    & 173.9   & 84.9  & 110.7    & 85.8   &  71.4   & 86.3   & \multicolumn{1}{c||}{73.1}     & {88.4}  \\ 
\multicolumn{1}{l||}{Chen et al \cite{chen_2d_match_cvpr17} }                              & 240.1            & 106.6       & 139.2       & 106.2       & 87.0  & 114.0   & \multicolumn{1}{c||}{90.5}         & 114.2       \\
\multicolumn{1}{l||}{Moreno et al \cite{moreno_distance_matrix_cvpr17} }                              & 116.7            & 87.7       & 100.4       & 94.6       & 75.2  & 87.8   & \multicolumn{1}{c||}{74.9}         & 85.6       \\
\multicolumn{1}{l||}{Zhou et al \cite{zhou2017towards} }                              & 111.6            & 64.1       & 65.5       &  66.0       & 63.2  & 51.4   & \multicolumn{1}{c||}{55.3}         & 64.9       \\
\multicolumn{1}{l||}{Martinez et al \cite{martinez20173dbaseline} }            & 94.6         & 62.3       & 78.4     &  59.1       & 49.5  & 65.1   & \multicolumn{1}{c||}{52.4}         & 62.9       \\
\multicolumn{1}{l||}{Tekin et al \cite{tekin_fusion_arxiv16} }            & 107.9         & 70.4       & 79.4    &  68.0       & 52.8  & 77.8   & \multicolumn{1}{c||}{63.1}         & 70.8       \\
\multicolumn{1}{l||}{Nie et al \cite{nie2017monocular} }            & 106.9         & 88.0       & 86.9    &  70.7       & 71.9  & 76.5   & \multicolumn{1}{c||}{73.2}         & 79.5       \\
\multicolumn{1}{l||}{VNect \cite{VNect_SIGGRAPH2017}\T}                              & 138.7            & 78.8       & 93.8       & 73.9       & {55.8}  & 82.0   & \multicolumn{1}{c||}{{59.6}}         & 80.5       \\
\multicolumn{1}{l||}{LCR-net \cite{rogez_lcr_cvpr17} }                              & 127.1            & 88.0       & 105.7       & 83.7       & 64.9  & 86.6   & \multicolumn{1}{c||}{84.0}         & 87.7       \\ \hline \hline
\multicolumn{1}{l||}{\T VNect (with our setup) }                                                                & 155.9       & 82.0   &  95.2   & 76.8   & 59.7   & 94.1    & \multicolumn{1}{c||}{64.3}     & {84.3}  \\ 
\multicolumn{1}{l||}{\T Our Single-Person}                                                                & {93.0}       & {68.8}   & {84.5}    & {65.1}   & 57.6   & {72.0}    & \multicolumn{1}{c||}{63.6}     & {{69.9}}  \\ 
\end{tabular}
}
\end{table*}

\begin{figure*}
  \begin{center}
  \includegraphics[width=1\linewidth]{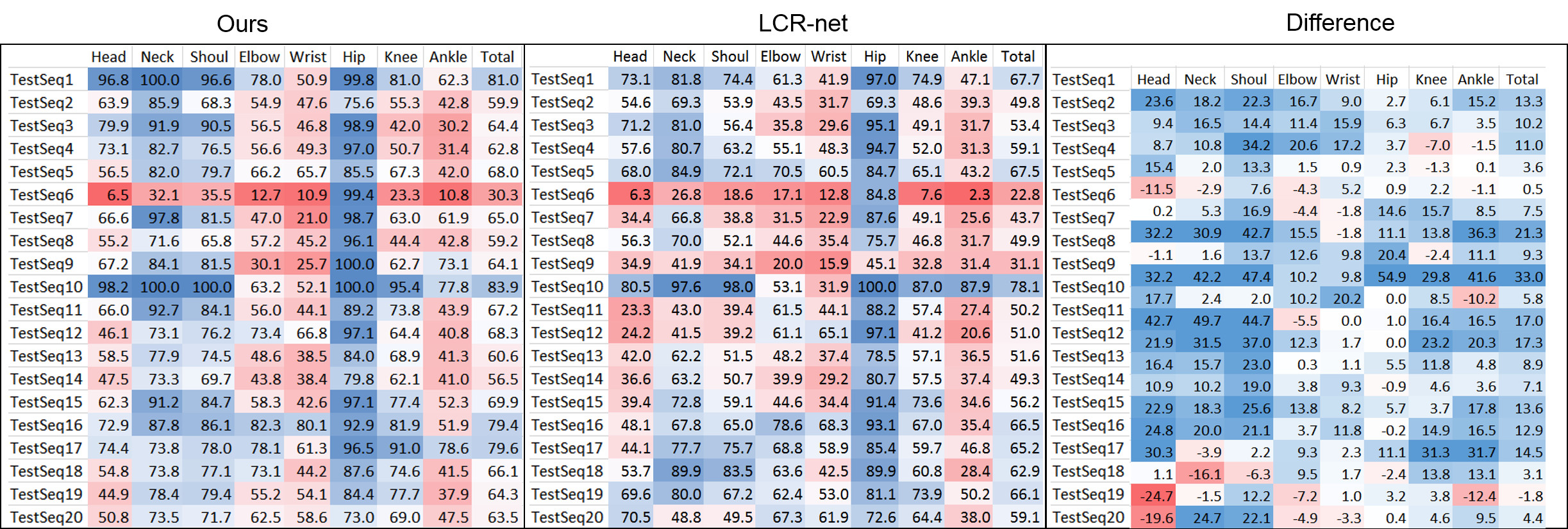}
  \end{center}
  \vspace{-0.3cm}
  \caption
  {Comparison of our method and LCR-net~\cite{rogez_lcr_cvpr17} on our proposed multi-person test set, here visualized as joint-wise breakdown of PCK for all 20 sequences, as well as the difference in accuracy between our method and LCR-net. LCR-net predictions were mapped to the ground truth bone lengths for fairness of comparison.}
  \vspace{-0.5cm}
  \label{fig:jointwise_multi}
\end{figure*}

\section{Evaluation on Single-person Test Sets}
Here we provide a detailed comparison against other methods for single-person 3D pose estimation. Evaluation on Human3.6m is in Table~\ref{tbl:s9_11_compare_full}, and on MPI-INF-3DHP test set in Table~\ref{tbl:our_testset_full}. 
We additionally provide comparisons with the VNect location-maps trained on our training setup, which includes the 2D pretraining, and the 3D pose samples.

\begin{table*}
\renewcommand{\tabcolsep}{1.5pt}
\centering
\caption{Comparison of our method against the state of the art on single person MPI-INF-3DHP test set. All evaluations use ground-truth bounding box crops around the subject. We report the \emph{Percentage of Correct Keypoints measure in 3D} (@150mm), and the Area Under the Curve for the same, as proposed by MPI-INF-3DHP. We additionally report the Mean Per Joint Position Error in mm. Higher PCK and AUC is better, and lower MPJPE is better.}
\label{tbl:our_testset_full}
\resizebox{1\linewidth}{!}{
\begin{tabular}{c||c|c|c|c|c|c|c||ccc}
                                                                          &   Stand/        &               & Sit On        & Crouch/       & On the        &               &               &                           & \multicolumn{1}{l}{}      & \multicolumn{1}{l}{} \\
Network   &  Walk          & Exercise      & Chair         & Reach         & Floor         & Sports        & Misc.         & \multicolumn{3}{c}{Total}                                                    \\ \hline
  & \T PCK           & PCK           & PCK           & PCK           & PCK           & PCK           & PCK           & \multicolumn{1}{c|}{PCK}  & \multicolumn{1}{c|}{AUC}  & MPJPE(mm)                \\ \hline \cline{2-11}
\multirow{1}{*}{\begin{tabular}[c]{@{}c@{}}VNect~\cite{VNect_SIGGRAPH2017}\end{tabular}}    \T & \textbf{87.7}          & \textbf{77.4}          & 74.7          & 72.9          & 51.3          & \textbf{83.3}          & \textbf{80.1}          & \multicolumn{1}{c|}{{76.6}} & \multicolumn{1}{c|}{{40.4}} & 124.7                \\ \hline
\multirow{1}{*}{\begin{tabular}[c]{@{}c@{}}LCR-net~\cite{rogez_lcr_cvpr17}\end{tabular}}                                               \T & 70.5          & {56.3} & {58.5} & {69.4} & {39.6} & 57.7          & 57.6          & \multicolumn{1}{c|}{59.7} & \multicolumn{1}{c|}{27.6} & 158.4                \\\hline
\multirow{1}{*}{\begin{tabular}[c]{@{}c@{}}Zhou et al.\cite{zhou2017towards}\end{tabular}}                                               \T & 85.4 & 71.0 & 60.7 & 71.4 & 37.8 & 70.9 & 74.4      & \multicolumn{1}{c|}{69.2} & \multicolumn{1}{c|}{32.5} & 137.1                \\\hline
\multirow{1}{*}{\begin{tabular}[c]{@{}c@{}}Mehta et al.\cite{mehta_mono_3dv17}\end{tabular}}                                               \T & 86.6          & {75.3} & {74.8} & {73.7} & {52.2} & 82.1          & 77.5          & \multicolumn{1}{c|}{75.7} & \multicolumn{1}{c|}{39.3} & 117.6                \\\hline \hline
\multirow{1}{*}{\begin{tabular}[c]{@{}c@{}}Ours Single-Person (Torso)\end{tabular}}    \T & {75.0}          & {64.8}          & 69.1          & 68.7          & 48.6          & {70.0}          & {60.6}          & \multicolumn{1}{c|}{{65.6}} & \multicolumn{1}{c|}{{32.6}} & 142.8                \\ \cline{2-11}
\multirow{1}{*}{\begin{tabular}[c]{@{}c@{}}Ours Single-Person (Full)\end{tabular}}\T                                                                               & 83.8          & 75.0          & \underline{77.8}          & \textbf{77.5}          & \underline{55.1}          & 80.4          & 72.5          & \multicolumn{1}{c|}{75.2} & \multicolumn{1}{c|}{37.8} & 122.2                \\ \hline 
\multirow{1}{*}{\begin{tabular}[c]{@{}c@{}}Ours Multi-Person (Torso)\end{tabular}}    \T & {73.7}          & {63.7}          & 64.6          & 65.8          & 44.7          & {69.5}          & {60.2}          & \multicolumn{1}{c|}{{63.6}} & \multicolumn{1}{c|}{{31.1}} & 146.8                \\ \cline{2-11}
\multirow{1}{*}{\begin{tabular}[c]{@{}c@{}}Ours Multi-Person (Full)\end{tabular}}\T                                                                               & 82.0          & 74.5          & {75.9}          & \underline{73.9}          & {51.6}          & 79.0          & 71.8          & \multicolumn{1}{c|}{73.4} & \multicolumn{1}{c|}{36.2} & 126.3                \\ \hline 
\multirow{1}{*}{\begin{tabular}[c]{@{}c@{}}VNect (our train. setup) \end{tabular}}    \T & {85.7}          & {75.4}          & \textbf{78.6}          & 72.3          & \textbf{60.2}          & {81.8}          & {73.4}          & \multicolumn{1}{c|}{{75.8}} & \multicolumn{1}{c|}{{38.9}} & 120.1                \\ \hline
\end{tabular}
}
\end{table*}

Table~\ref{tbl:our_testset_occ} provides a sequencewise breakdown for the synthetic occlusion experiment on MPI-INF-3DHP test set wherein through randomly placed occlusions $\approx$14\% of the joints are occluded. This doesn't account for self-occlusions.
\begin{table*}
\centering
\caption{Testing occlusion robustness of our method through synthetic occlusions on MPI-INF-3DHP single person test set. The synthetic occlusions cover about 14\% of the evaluated joints overall. We report the \emph{Percentage of Correct Keypoints measure in 3D} (@150mm) overall, as well as split by occlusion. Higher PCK.}
\label{tbl:our_testset_occ}
\resizebox{0.8\linewidth}{!}{
\begin{tabular}{c||c|c|c|c|c|c||c}
   &  Seq1          & Seq2      & Seq3         & Seq4         & Seq5         & Seq6           & \multicolumn{1}{c}{Total}                                                    \\ \cline{2-8}
  & \T PCK           & PCK           & PCK           & PCK           & PCK           & PCK           & PCK                           \\ \hline \cline{2-8}
\multicolumn{8}{l}{ \textbf{Overall}\T } \\ \hline
\T \multirow{1}{*}{\begin{tabular}[c]{@{}c@{}}Ours Multi-Person\end{tabular}}                                                                               & 78.7          & 70.0          & 71.9          & 65.2          & 61.4          & 60.7          & 69.0                        \\ 
\multirow{1}{*}{\begin{tabular}[c]{@{}c@{}}Ours Single-Person \end{tabular}}                                                                               & 80.9          & 72.8          & 72.6          & 65.7         & 62.5          &  65.8         & 71.1                         \\ 
\multirow{1}{*}{\begin{tabular}[c]{@{}c@{}}VNect \cite{VNect_SIGGRAPH2017}\end{tabular}}                                                                               & 80.1          & 72.4          & 72.4          & 61.5          & 50.2          & 69.8          & 69.4                          \\ 
\multirow{1}{*}{\begin{tabular}[c]{@{}c@{}}VNect (our train. setup)\end{tabular}}                                                                               & 79.3          & 74.4          & 72.2          &  67.2          & 55.7          & 64.6          & 70.4                          \\ \hline 

\multicolumn{8}{l}{ \textbf{Occluded Subset of Joints}\T } \\ \hline
\T \multirow{1}{*}{\begin{tabular}[c]{@{}c@{}}Ours Multi-Person\end{tabular}}                                                                               & 73.3          & 66.5          &  55.0          & 56.5          &  45.1          & 64.9          & 62.8                        \\ 
\multirow{1}{*}{\begin{tabular}[c]{@{}c@{}}Ours Single-Person \end{tabular}}                                                                               & 74.9          & 63.2          &  59.0        & 54.2          &  48.0          & 68.4          & 64.0                         \\ 
\multirow{1}{*}{\begin{tabular}[c]{@{}c@{}}VNect \cite{VNect_SIGGRAPH2017}\end{tabular}}                                                                               & 61.4          & 54.5          & 47.6          & 36.4          & 30.5          & 66.2          & 53.2                          \\ 
\multirow{1}{*}{\begin{tabular}[c]{@{}c@{}}VNect (our train. setup)\end{tabular}}                                                                               & 69.6          & 61.9          & {49.0}          & 50.8          & 43.5          & 63.4          & 59.2                          \\ \hline 

\multicolumn{8}{l}{ \textbf{Un-occluded Subset of Joints}\T } \\ \hline
\T \multirow{1}{*}{\begin{tabular}[c]{@{}c@{}}Ours Multi-Person\end{tabular}}                                                                               & 79.9          & 70.5          & 73.7          & 66.2          & 64.6          & 59.5          & 70.0                        \\ 
\multirow{1}{*}{\begin{tabular}[c]{@{}c@{}}Ours Single-Person \end{tabular}}                                                                               & 82.1          & 74.0          & 74.1          &  67.0         &  65.3          & 65.1          & 72.2                         \\ 
\multirow{1}{*}{\begin{tabular}[c]{@{}c@{}}VNect \cite{VNect_SIGGRAPH2017}\end{tabular}}                                                                               & 83.9          & 74.6          & {75.0}          & 64.4          & 54.0          & 70.9          & 72.1                          \\ 
\multirow{1}{*}{\begin{tabular}[c]{@{}c@{}}VNect (our train. setup)\end{tabular}}                                                                               & 81.3          & 76.0          & 74.6          & 69.0          & 58.1          & 64.8          & 72.2                          \\ \hline 
\end{tabular}
}
\vspace{-0.5cm}
\end{table*}

\begin{figure*}
\centering
  \includegraphics[width=0.88\linewidth]{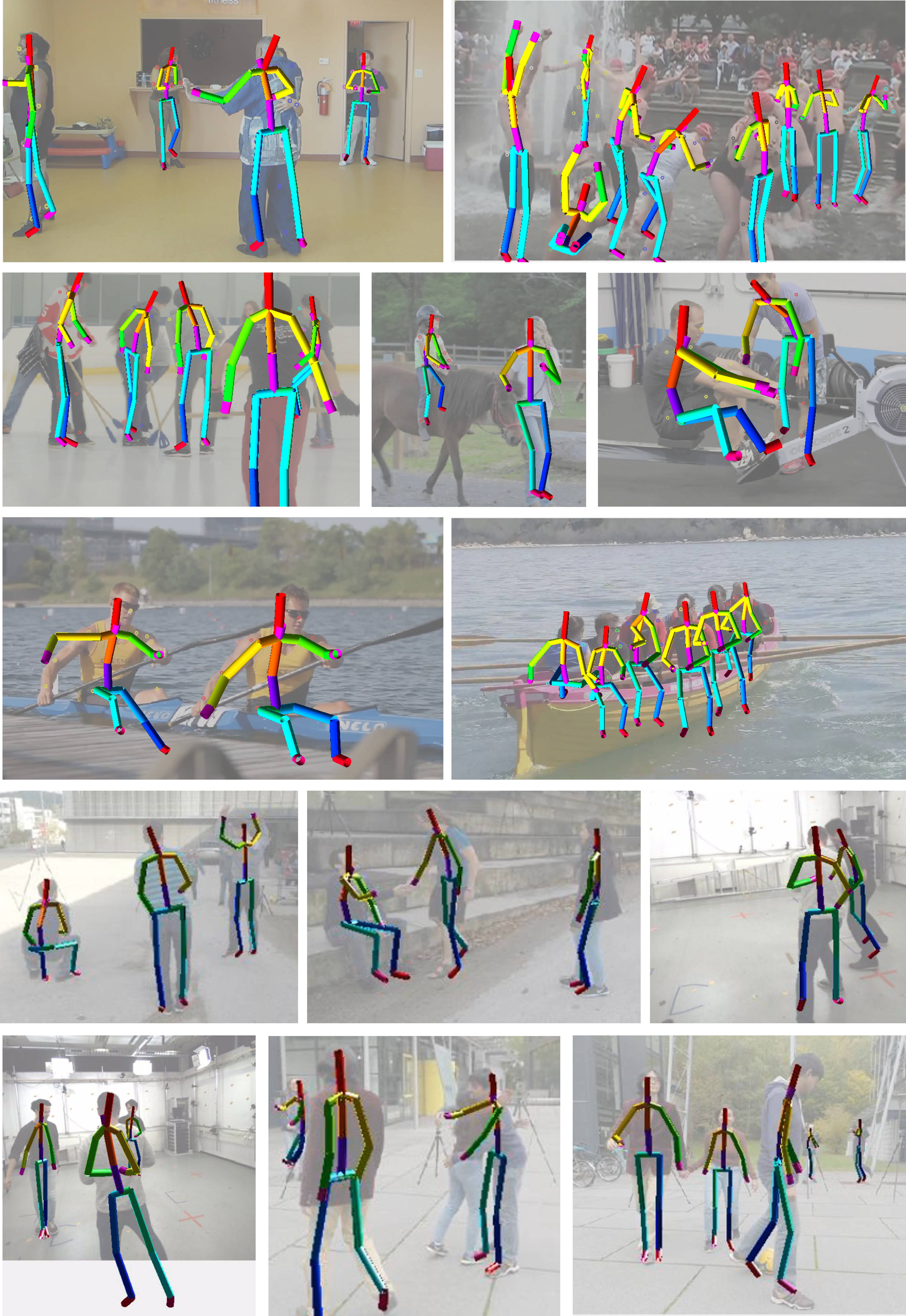}
  \vspace{-0.0cm}
  \caption
  {More qualitative results of our approach on MPI 2D pose dataset~\cite{andriluka_mpii2d_cvpr14} and our proposed MuPoTS-3D test set.
  }  
  \label{fig:qualitative_results}
  \vspace{-0.4cm}
\end{figure*}

%
%

\clearpage
\clearpage
{\small
\bibliographystyle{ieee}
\bibliography{article}

\begin{thebibliography}{10}\itemsep=-1pt

\bibitem{akhter_pose_conditioned_cvpr15}
I.~Akhter and M.~J. Black.
\newblock Pose-conditioned joint angle limits for 3d human pose reconstruction.
\newblock In {\em IEEE Conference on Computer Vision and Pattern Recognition
  (CVPR)}, pages 1446--1455, 2015.

\bibitem{alldieck2018video}
T.~Alldieck, M.~Magnor, W.~Xu, C.~Theobalt, and G.~Pons-Moll.
\newblock Video based reconstruction of 3d people models.
\newblock In {\em {IEEE} Conf. on Computer Vision and Pattern Recognition},
  2018.

\bibitem{andriluka_mpii2d_cvpr14}
M.~Andriluka, L.~Pishchulin, P.~Gehler, and B.~Schiele.
\newblock {2D Human Pose Estimation: New Benchmark and State of the Art
  Analysis}.
\newblock In {\em IEEE Conference on Computer Vision and Pattern Recognition
  (CVPR)}, June 2014.

\bibitem{andriluka_pictorial_cvpr09}
M.~Andriluka, S.~Roth, and B.~Schiele.
\newblock Pictorial structures revisited: People detection and articulated pose
  estimation.
\newblock In {\em IEEE Conference on Computer Vision and Pattern Recognition
  (CVPR)}, pages 1014--1021, 2009.

\bibitem{bo_twin_ijcv10}
L.~Bo and C.~Sminchisescu.
\newblock Twin gaussian processes for structured prediction.
\newblock In {\em International Journal of Computer Vision}, 2010.

\bibitem{bogo_smpl_eccv16}
F.~Bogo, A.~Kanazawa, C.~Lassner, P.~Gehler, J.~Romero, and M.~J. Black.
\newblock Keep it {SMPL}: Automatic estimation of {3D} human pose and shape
  from a single image.
\newblock In {\em European Conference on Computer Vision (ECCV)}, 2016.

\bibitem{bulat_convpart_eccv16}
A.~Bulat and G.~Tzimiropoulos.
\newblock Human pose estimation via convolutional part heatmap regression.
\newblock In {\em European Conference on Computer Vision (ECCV)}, 2016.

\bibitem{cao_affinity_2017}
Z.~Cao, T.~Simon, S.~Wei, and Y.~Sheikh.
\newblock Realtime multi-person 2d pose estimation using part affinity fields.
\newblock In {\em IEEE Conference on Computer Vision and Pattern Recognition
  (CVPR)}, 2017.

\bibitem{chen_2d_match_cvpr17}
C.-H. Chen and D.~Ramanan.
\newblock 3d human pose estimation = 2d pose estimation + matching.
\newblock In {\em CVPR 2017-IEEE Conference on Computer Vision \& Pattern
  Recognition}, 2017.

\bibitem{chen_synth_data_3dv16}
W.~Chen, H.~Wang, Y.~Li, H.~Su, Z.~Wang, C.~Tu, D.~Lischinski, D.~Cohen-Or, and
  B.~Chen.
\newblock Synthesizing training images for boosting human 3d pose estimation.
\newblock In {\em International Conference on 3D Vision (3DV)}, 2016.

\bibitem{chen_nips14}
X.~Chen and A.~L. Yuille.
\newblock Articulated pose estimation by a graphical model with image dependent
  pairwise relations.
\newblock In {\em Advances in Neural Information Processing Systems (NIPS)},
  pages 1736--1744, 2014.

\bibitem{elhayek_convmocap_TPAMI2016}
A.~Elhayek, E.~Aguiar, A.~Jain, J.~Tompson, L.~Pishchulin, M.~Andriluka,
  C.~Bregler, B.~Schiele, and C.~Theobalt.
\newblock {MARCOnI - ConvNet-based MARker-less Motion Capture in Outdoor and
  Indoor Scenes}.
\newblock {\em IEEE Transactions on Pattern Analysis and Machine Intelligence
  (PAMI)}, 2016.

\bibitem{gkioxari2014using}
G.~Gkioxari, B.~Hariharan, R.~Girshick, and J.~Malik.
\newblock Using k-poselets for detecting people and localizing their keypoints.
\newblock In {\em Proceedings of the IEEE Conference on Computer Vision and
  Pattern Recognition}, pages 3582--3589, 2014.

\bibitem{guler2018densepose}
R.~A. G{\"u}ler, N.~Neverova, and I.~Kokkinos.
\newblock Densepose: Dense human pose estimation in the wild.

\bibitem{he_resnet_cvpr2016}
K.~He, X.~Zhang, S.~Ren, and J.~Sun.
\newblock Deep residual learning for image recognition.
\newblock In {\em IEEE Conference on Computer Vision and Pattern Recognition
  (CVPR)}, 2016.

\bibitem{insafutdin_arttrack_cvpr17}
E.~Insafutdinov, M.~Andriluka, L.~Pishchulin, S.~Tang, E.~Levinkov, B.~Andres,
  B.~Schiele, and S.~I. Campus.
\newblock Arttrack: Articulated multi-person tracking in the wild.
\newblock In {\em Proc. of CVPR}, 2017.

\bibitem{insafutdinov_deepercut_eccv16}
E.~Insafutdinov, L.~Pishchulin, B.~Andres, M.~Andriluka, and B.~Schiele.
\newblock Deepercut: A deeper, stronger, and faster multi-person pose
  estimation model.
\newblock In {\em European Conference on Computer Vision (ECCV)}, 2016.

\bibitem{ionescu_human36_pami14}
C.~Ionescu, D.~Papava, V.~Olaru, and C.~Sminchisescu.
\newblock {Human3.6m: Large scale datasets and predictive methods for 3d human
  sensing in natural environments}.
\newblock {\em IEEE Transactions on Pattern Analysis and Machine Intelligence
  (PAMI)}, 36(7):1325--1339, 2014.

\bibitem{iqbal2016multi}
U.~Iqbal and J.~Gall.
\newblock Multi-person pose estimation with local joint-to-person associations.
\newblock In {\em European Conference on Computer Vision Workshops}, pages
  627--642. Springer, 2016.

\bibitem{Jahangiri2017}
E.~Jahangiri and A.~L. Yuille.
\newblock Generating multiple diverse hypotheses for human 3d pose consistent
  with 2d joint detections.
\newblock In {\em IEEE International Conference on Computer Vision (ICCV)
  Workshops (PeopleCap)}, 2017.

\bibitem{jia_caffe_ICM}
Y.~Jia, E.~Shelhamer, J.~Donahue, S.~Karayev, J.~Long, R.~Girshick,
  S.~Guadarrama, and T.~Darrell.
\newblock Caffe: Convolutional architecture for fast feature embedding.
\newblock In {\em Proceedings of the 22nd ACM International Conference on
  Multimedia}, pages 675--678, 2014.

\bibitem{johnson_lsp_bmvc10}
S.~Johnson and M.~Everingham.
\newblock Clustered pose and nonlinear appearance models for human pose
  estimation.
\newblock In {\em British Machine Vision Conference (BMVC)}, 2010.
\newblock doi:10.5244/C.24.12.

\bibitem{johnson_lspet_cvpr11}
S.~Johnson and M.~Everingham.
\newblock Learning effective human pose estimation from inaccurate annotation.
\newblock In {\em Proceedings of IEEE Conference on Computer Vision and Pattern
  Recognition}, 2011.

\bibitem{joo_panoptic_iccv2015}
H.~Joo, H.~Liu, L.~Tan, L.~Gui, B.~Nabbe, I.~Matthews, T.~Kanade, S.~Nobuhara,
  and Y.~Sheikh.
\newblock Panoptic studio: A massively multiview system for social motion
  capture.
\newblock In {\em ICCV}, pages 3334--3342, 2015.

\bibitem{hmrKanazawa17}
A.~Kanazawa, M.~J. Black, D.~W. Jacobs, and J.~Malik.
\newblock End-to-end recovery of human shape and pose.
\newblock In {\em Computer Vision and Pattern Regognition (CVPR)}, 2018.

\bibitem{Lassner:UP:2017}
C.~Lassner, J.~Romero, M.~Kiefel, F.~Bogo, M.~J. Black, and P.~V. Gehler.
\newblock Unite the people: Closing the loop between 3d and 2d human
  representations.
\newblock In {\em IEEE Conf. on Computer Vision and Pattern Recognition
  (CVPR)}, July 2017.

\bibitem{li_accv14}
S.~Li and A.~B. Chan.
\newblock 3d human pose estimation from monocular images with deep
  convolutional neural network.
\newblock In {\em Asian Conference on Computer Vision (ACCV)}, pages 332--347,
  2014.

\bibitem{li_maximum_iccv2015}
S.~Li, W.~Zhang, and A.~B. Chan.
\newblock Maximum-margin structured learning with deep networks for 3d human
  pose estimation.
\newblock In {\em IEEE International Conference on Computer Vision (ICCV)},
  pages 2848--2856, 2015.

\bibitem{lin_coco_eccv14}
T.-Y. Lin, M.~Maire, S.~Belongie, J.~Hays, P.~Perona, D.~Ramanan,
  P.~Doll{\'a}r, and C.~L. Zitnick.
\newblock Microsoft coco: Common objects in context.
\newblock In {\em European conference on computer vision}, pages 740--755.
  Springer, 2014.

\bibitem{SMPL:2015}
M.~Loper, N.~Mahmood, J.~Romero, G.~Pons-Moll, and M.~J. Black.
\newblock {SMPL}: A skinned multi-person linear model.
\newblock {\em ACM Trans. Graphics (Proc. SIGGRAPH Asia)}, 34(6):248:1--248:16,
  Oct. 2015.

\bibitem{luvizon20182d}
D.~C. Luvizon, D.~Picard, and H.~Tabia.
\newblock 2d/3d pose estimation and action recognition using multitask deep
  learning.
\newblock In {\em The IEEE Conference on Computer Vision and Pattern
  Recognition (CVPR)}, volume~2, 2018.

\bibitem{martinez20173dbaseline}
J.~Martinez, R.~Hossain, J.~Romero, and J.~J. Little.
\newblock A simple yet effective baseline for 3d human pose estimation.
\newblock In {\em IEEE International Conference on Computer Vision (ICCV)},
  2017.

\bibitem{mehta_mono_3dv17}
D.~Mehta, H.~Rhodin, D.~Casas, P.~Fua, O.~Sotnychenko, W.~Xu, and C.~Theobalt.
\newblock Monocular 3d human pose estimation in the wild using improved cnn
  supervision.
\newblock In {\em 3D Vision (3DV), 2017 Fifth International Conference on},
  2017.

\bibitem{VNect_SIGGRAPH2017}
D.~Mehta, S.~Sridhar, O.~Sotnychenko, H.~Rhodin, M.~Shafiei, H.-P. Seidel,
  W.~Xu, D.~Casas, and C.~Theobalt.
\newblock Vnect: Real-time 3d human pose estimation with a single rgb camera.
\newblock volume~36, 2017.

\bibitem{moreno_distance_matrix_cvpr17}
F.~Moreno-Noguer.
\newblock 3d human pose estimation from a single image via distance matrix
  regression.
\newblock In {\em CVPR 2017-IEEE Conference on Computer Vision \& Pattern
  Recognition}, 2017.

\bibitem{newell_associative_nips17}
A.~Newell and J.~Deng.
\newblock Associative embedding: End-to-end learning for joint detection and
  grouping.
\newblock In {\em Advances in Neural Information Processing Systems (NIPS)},
  2017.

\bibitem{newell_stacked_hourglass_eccv16}
A.~Newell, K.~Yang, and J.~Deng.
\newblock Stacked hourglass networks for human pose estimation.
\newblock In {\em European Conference on Computer Vision (ECCV)}, 2016.

\bibitem{nie2017monocular}
B.~X. Nie, P.~Wei, and S.-C. Zhu.
\newblock Monocular 3d human pose estimation by predicting depth on joints.
\newblock In {\em IEEE International Conference on Computer Vision}, 2017.

\bibitem{omran2018neural}
M.~Omran, C.~Lassner, G.~Pons-Moll, P.~Gehler, and B.~Schiele.
\newblock Neural body fitting: Unifying deep learning and model based human
  pose and shape estimation.
\newblock In {\em International Conf. on 3D Vision}, 2018.

\bibitem{papandreou2017towards}
G.~Papandreou, T.~Zhu, N.~Kanazawa, A.~Toshev, J.~Tompson, C.~Bregler, and
  K.~Murphy.
\newblock Towards accurate multi-person pose estimation in the wild.
\newblock {\em arXiv preprint arXiv:1701.01779}, 2017.

\bibitem{pavlakos_volumetric_cvpr17}
G.~Pavlakos, X.~Zhou, K.~G. Derpanis, and K.~Daniilidis.
\newblock Coarse-to-fine volumetric prediction for single-image 3{D} human
  pose.
\newblock In {\em CVPR 2017-IEEE Conference on Computer Vision \& Pattern
  Recognition}, 2017.

\bibitem{pavlakos2018humanshape}
G.~Pavlakos, L.~Zhu, X.~Zhou, and K.~Daniilidis.
\newblock Learning to estimate 3{D} human pose and shape from a single color
  image.
\newblock In {\em CVPR}, 2018.

\bibitem{pishchulin_deepcut_cvpr16}
L.~Pishchulin, E.~Insafutdinov, S.~Tang, B.~Andres, M.~Andriluka, P.~Gehler,
  and B.~Schiele.
\newblock Deepcut: Joint subset partition and labeling for multi person pose
  estimation.
\newblock In {\em IEEE Conference on Computer Vision and Pattern Recognition
  (CVPR)}, 2016.

\bibitem{pishchulin_reshape_cvpr12}
L.~Pishchulin, A.~Jain, M.~Andriluka, T.~Thorm{\"a}hlen, and B.~Schiele.
\newblock Articulated people detection and pose estimation: Reshaping the
  future.
\newblock In {\em Conference on Computer Vision and Pattern Recognition
  (CVPR)}, pages 3178--3185. IEEE, 2012.

\bibitem{posebits_cvpr14}
G.~Pons-Moll, D.~J. Fleet, and B.~Rosenhahn.
\newblock Posebits for monocular human pose estimation.
\newblock In {\em IEEE Conference on Computer Vision and Pattern Recognition
  (CVPR)}, pages 2337--2344, 2014.

\bibitem{popa2017deep}
A.-I. Popa, M.~Zanfir, and C.~Sminchisescu.
\newblock Deep multitask architecture for integrated 2d and 3d human sensing.
\newblock {\em IEEE Conference on Computer Vision and Pattern Recognition
  (CVPR)}, 2017.

\bibitem{ren_faster_rcnn_nips15}
S.~Ren, K.~He, R.~Girshick, and J.~Sun.
\newblock Faster r-cnn: Towards real-time object detection with region proposal
  networks.
\newblock In {\em Advances in neural information processing systems}, pages
  91--99, 2015.

\bibitem{rogez_mocap_nips16}
G.~Rogez and C.~Schmid.
\newblock Mocap-guided data augmentation for 3d pose estimation in the wild.
\newblock In {\em Advances in Neural Information Processing Systems}, pages
  3108--3116, 2016.

\bibitem{rogez_lcr_cvpr17}
G.~Rogez, P.~Weinzaepfel, and C.~Schmid.
\newblock Lcr-net: Localization-classification-regression for human pose.
\newblock In {\em CVPR 2017-IEEE Conference on Computer Vision \& Pattern
  Recognition}, 2017.

\bibitem{sarafianos_posesurvey_cviu2016}
N.~Sarafianos, B.~Boteanu, B.~Ionescu, and I.~A. Kakadiaris.
\newblock 3d human pose estimation: A review of the literature and analysis of
  covariates.
\newblock {\em Computer Vision and Image Understanding}, 152:1--20, 2016.

\bibitem{sigal_humaneva_ijcv10}
L.~Sigal, A.~O. Balan, and M.~J. Black.
\newblock Humaneva: Synchronized video and motion capture dataset and baseline
  algorithm for evaluation of articulated human motion.
\newblock {\em International Journal of Computer Vision (IJCV)}, 87(1-2):4--27,
  2010.

\bibitem{simo_joint_CVPR2013}
E.~Simo-Serra, A.~Quattoni, C.~Torras, and F.~Moreno-Noguer.
\newblock A joint model for 2d and 3d pose estimation from a single image.
\newblock In {\em Conference on Computer Vision and Pattern Recognition
  (CVPR)}, pages 3634--3641, 2013.

\bibitem{simon2017hand}
T.~Simon, H.~Joo, I.~Matthews, and Y.~Sheikh.
\newblock Hand keypoint detection in single images using multiview
  bootstrapping.
\newblock In {\em IEEE Conference on Computer Vision and Pattern Recognition
  (CVPR)}, 2017.

\bibitem{sminchisescu2003kinematic}
C.~Sminchisescu and B.~Triggs.
\newblock Kinematic jump processes for monocular 3d human tracking.
\newblock In {\em Computer Vision and Pattern Recognition, 2003. Proceedings.
  2003 IEEE Computer Society Conference on}, volume~1, pages I--I. IEEE, 2003.

\bibitem{sun2011articulated}
M.~Sun and S.~Savarese.
\newblock Articulated part-based model for joint object detection and pose
  estimation.
\newblock In {\em IEEE International Conference on Computer Vision}, pages
  723--730. IEEE, 2011.

\bibitem{sun2017compositional}
X.~Sun, J.~Shang, S.~Liang, and Y.~Wei.
\newblock Compositional human pose regression.
\newblock {\em arXiv preprint arXiv:1704.00159}, 2017.

\bibitem{tan2017indirect}
J.~K.~V. Tan, I.~Budvytis, and R.~Cipolla.
\newblock Indirect deep structured learning for 3d human body shape and pose
  prediction.
\newblock In {\em BMVC}, volume~3, page~6, 2017.

\bibitem{taylor_articulated_cvpr00}
C.~J. Taylor.
\newblock Reconstruction of articulated objects from point correspondences in a
  single uncalibrated image.
\newblock In {\em IEEE Conference on Computer Vision and Pattern Recognition
  (CVPR)}, volume~1, pages 677--684, 2000.

\bibitem{tekin_structured_bmvc16}
B.~Tekin, I.~Katircioglu, M.~Salzmann, V.~Lepetit, and P.~Fua.
\newblock {Structured Prediction of 3D Human Pose with Deep Neural Networks}.
\newblock In {\em British Machine Vision Conference (BMVC)}, 2016.

\bibitem{tekin_fusion_arxiv16}
B.~{Tekin}, P.~{M{\'a}rquez-Neila}, M.~{Salzmann}, and P.~{Fua}.
\newblock Learning to fuse 2d and 3d image cues for monocular body pose
  estimation.
\newblock In {\em IEEE International Conference on Computer Vision (ICCV)},
  2017.

\bibitem{Captury}
{The Captury}.
\newblock \url{http://www.thecaptury.com/}, 2016.

\bibitem{tome_lifting_2017}
D.~Tome, C.~Russell, and L.~Agapito.
\newblock Lifting from the deep: Convolutional 3d pose estimation from a single
  image.
\newblock In {\em IEEE Conference on Computer Vision and Pattern Recognition
  (CVPR)}, 2017.

\bibitem{trumble2017total}
M.~Trumble, A.~Gilbert, C.~Malleson, A.~Hilton, and J.~Collomosse.
\newblock Total capture: 3d human pose estimation fusing video and inertial
  sensors.
\newblock In {\em Proceedings of 28th British Machine Vision Conference}, pages
  1--13, 2017.

\bibitem{vonPon2016a}
T.~von Marcard, G.~Pons-Moll, and B.~Rosenhahn.
\newblock Human pose estimation from video and imus.
\newblock {\em Transactions on Pattern Analysis and Machine Intelligence},
  38(8):1533--1547, Jan. 2016.

\bibitem{yang20183d}
W.~Yang, W.~Ouyang, X.~Wang, J.~Ren, H.~Li, and X.~Wang.
\newblock 3d human pose estimation in the wild by adversarial learning.
\newblock In {\em Proceedings of the IEEE Conference on Computer Vision and
  Pattern Recognition}, volume~1, 2018.

\bibitem{yasin_dual_source_cvpr16}
H.~Yasin, U.~Iqbal, B.~Kr{\"u}ger, A.~Weber, and J.~Gall.
\newblock {A Dual-Source Approach for 3D Pose Estimation from a Single Image}.
\newblock In {\em Conference on Computer Vision and Pattern Recognition
  (CVPR)}, 2016.

\bibitem{zanfir2018monocular}
A.~Zanfir, E.~Marinoiu, and C.~Sminchisescu.
\newblock Monocular 3d pose and shape estimation of multiple people in natural
  scenes--the importance of multiple scene constraints.
\newblock In {\em Proceedings of the IEEE Conference on Computer Vision and
  Pattern Recognition}, pages 2148--2157, 2018.

\bibitem{zhou2017towards}
X.~Zhou, Q.~Huang, X.~Sun, X.~Xue, and Y.~Wei.
\newblock Towards 3d human pose estimation in the wild: A weakly-supervised
  approach.
\newblock In {\em Proceedings of the IEEE Conference on Computer Vision and
  Pattern Recognition}, pages 398--407, 2017.

\bibitem{zhou_sparseness_deepness_cvpr15}
X.~Zhou, M.~Zhu, S.~Leonardos, K.~Derpanis, and K.~Daniilidis.
\newblock {Sparseness Meets Deepness: 3D Human Pose Estimation from Monocular
  Video}.
\newblock In {\em IEEE Conference on Computer Vision and Pattern Recognition
  (CVPR)}, 2015.

\end{thebibliography}
}

\end{document}